\documentclass{article}

\usepackage[preprint]{neurips_2025}

\usepackage[utf8]{inputenc} 
\usepackage[T1]{fontenc}    
\usepackage{hyperref}       
\usepackage{url}            
\usepackage{booktabs}       
\usepackage{amsfonts}       
\usepackage{nicefrac}       
\usepackage{microtype}      
\usepackage{xcolor}         
\usepackage{amsmath,amssymb,amsthm}
\usepackage[capitalize]{cleveref}
\usepackage{graphicx}
\usepackage{multirow}
\usepackage{makecell}
\usepackage{listings}
\usepackage{enumitem}
\usepackage{wrapfig}
\usepackage{subcaption}
\usepackage{xspace}

\usepackage[skins,theorems]{tcolorbox}

\tcbset{
  aibox/.style={
    width=\textwidth,
    top=7pt,
    bottom=5pt,
    colback=blue!6!white,
    colframe=black,
    colbacktitle=black,
    enhanced,
    center,
    attach boxed title to top left={yshift=-0.1in,xshift=0.15in},
    boxed title style={boxrule=0pt,colframe=white,},
  }
}
\newtcolorbox{AIbox}[2][]{aibox,title=#2,#1}

\crefname{section}{§\!}{§\!}

\newtheorem{remark}{Remark}

\definecolor{mygreen}{RGB}{117,159,96}
\definecolor{myred}{RGB}{206,109,109}
\definecolor{myyellow}{RGB}{201,201,69}
\definecolor{jtlred}{RGB}{224,108,117}

\newcommand{\framework}[0]{\texttt{DIET}\xspace}

\title{The Overthinker's \underline{DIET}: Cutting Token Calories with \underline{DI}fficulty\mbox{-}Awar\underline{E} \underline{T}raining}

\author{%
  Weize Chen\thanks{Equal Contributions}, Jiarui Yuan\footnotemark[1], Tailin Jin, Ning Ding, Huimin Chen, Zhiyuan Liu\thanks{Corresponding Author}, Maosong Sun\\
  Tsinghua University\\
  \texttt{\{chenwz21, yuanjr22\}@mails.tsinghua.edu.cn, liuzy@tsinghua.edu.cn}
}

\begin{document}

\maketitle

\begin{abstract}
    Recent large language models (LLMs) exhibit impressive reasoning but often \textit{overthink}, generating excessively long responses that hinder efficiency. We introduce \textbf{\framework (\underline{DI}fficulty-Awar\underline{E} \underline{T}raining)}, a framework that systematically cuts these "token calories" by integrating on-the-fly problem difficulty into the reinforcement learning (RL) process. \framework dynamically adapts token compression strategies by modulating token penalty strength and conditioning target lengths on estimated task difficulty, to optimize the performance-efficiency trade-off. We also theoretically analyze the pitfalls of naive reward weighting in group-normalized RL algorithms like GRPO, and propose \textit{Advantage Weighting} technique, which enables stable and effective implementation of these difficulty-aware objectives.
    Experimental results demonstrate that \framework significantly reduces token counts while simultaneously \textit{improving} reasoning performance. Beyond raw token reduction, we show two crucial benefits largely overlooked by prior work: (1) \framework leads to superior \textbf{inference scaling}. By maintaining high per-sample quality with fewer tokens, it enables better scaling performance via majority voting with more samples under fixed computational budgets, an area where other methods falter. (2) \framework enhances the natural positive correlation between response length and problem difficulty, ensuring verbosity is appropriately allocated, unlike many existing compression methods that disrupt this relationship. Our analyses provide a principled and effective framework for developing more efficient, practical, and high-performing LLMs.
\end{abstract}

\section{Introduction}
\label{sec:introduction}

Recent breakthroughs in large language models (LLMs) have yielded remarkable reasoning capabilities, particularly when enhanced through reinforcement learning (RL) from outcome-based rewards~\citep{openai2024learning,DBLP:journals/corr/abs-2501-12948,qwq32b}. These models excel in complex domains like mathematics and coding, often generating sophisticated reasoning chains~\citep{gandhi2025cognitive,tinyzero,deepscaler2025}. However, this enhanced reasoning frequently comes with a significant side effect: a dramatic increase in response length compared to base or instruction-tuned models. While some verbosity can facilitate complex thought, it often leads to \textbf{\textit{overthinking}}: models produce excessively long responses, sometimes thousands of tokens, even for simple queries (e.g., "2+3=?")~\citep{DBLP:journals/corr/abs-2412-21187,DBLP:journals/corr/abs-2502-03373,DBLP:journals/corr/abs-2501-12570}. This verbosity severely impacts inference latency and computational costs, hindering the practical deployment of these powerful reasoning models. Initial attempts to mitigate overthinking via supervised fine-tuning (SFT), direct preference optimization (DPO), or simple length penalties in RL objectives~\citep{team2025kimi,DBLP:journals/corr/abs-2502-12067,aggarwal2025l1} often struggle, leading to performance degradation, i.e., models that are concise but inaccurate.

A crucial dimension often overlooked in token compression is the intrinsic link between \textbf{problem difficulty} and the \textit{appropriate} level of verbosity. We contend that a "one-size-fits-all" compression strategy is fundamentally flawed. Complex problems may necessitate longer, more detailed reasoning, whereas simpler ones should elicit direct, concise answers. Indeed, as we later show (\cref{sec:preliminary-corr-analysis}), LLMs often naturally exhibit a tendency to use more tokens for problems they find more challenging. Token compression methods that ignore this inherent relationship could force a suboptimal compromise across the difficulty spectrum. We argue that explicitly incorporating on-the-fly difficulty estimation is essential for developing token compression strategies that effectively preserve reasoning capabilities.

To address these challenges, we introduce \textbf{\framework (\underline{DI}fficulty-Awar\underline{E} \underline{T}raining)}, a framework to systematically "cut token calories" from overthinking LLMs. \framework integrates on-the-fly problem difficulty into the RL training process, dynamically adapting token compression based on estimated task difficulty. This enables \framework to selectively encourage conciseness for simpler problems while preserving necessary verbosity for complex ones, improving the performance-efficiency trade-off and achieving better reasoning with significantly fewer tokens than the base model.

The effective implementation of such difficulty-aware objectives within popular RL algorithms like GRPO~\citep{DBLP:journals/corr/abs-2402-03300} presents its own challenges. We theoretically analyze the pitfalls of naively applying weighted rewards in these settings and propose a robust \textit{Advantage Weighting} technique that ensures stable and effective token compression training. Beyond standard performance gains, our work reveals two often-overlooked benefits of difficulty-aware compression:
\textbf{(1)} Critically for practical deployment, we show that \framework leads to superior \textbf{inference scaling}. By maintaining high per-sample quality with fewer tokens, it enables significantly better inference scaling performance under small computational budgets, an area where most prior token compression efforts falter or show degradation.
\textbf{(2)} We demonstrate that \framework not only compresses, but also \textit{enhances} the natural positive correlation between an LLM's response length and problem difficulty: a desirable characteristic for adaptive reasoning that many existing compression methods inadvertently disrupt.

This paper thus formalizes the difficulty-aware token compression problem and presents \framework as a comprehensive solution. Through the core adaptive mechanisms, the enabling Advantage Weighting technique, and by demonstrating unique benefits in enhancing inference scaling and preserving adaptive verbosity, \framework offers a principled and highly effective approach for developing more efficient, practical, and powerful reasoning LLMs.
\begin{figure}
    \centering
    \includegraphics[width=0.98\linewidth]{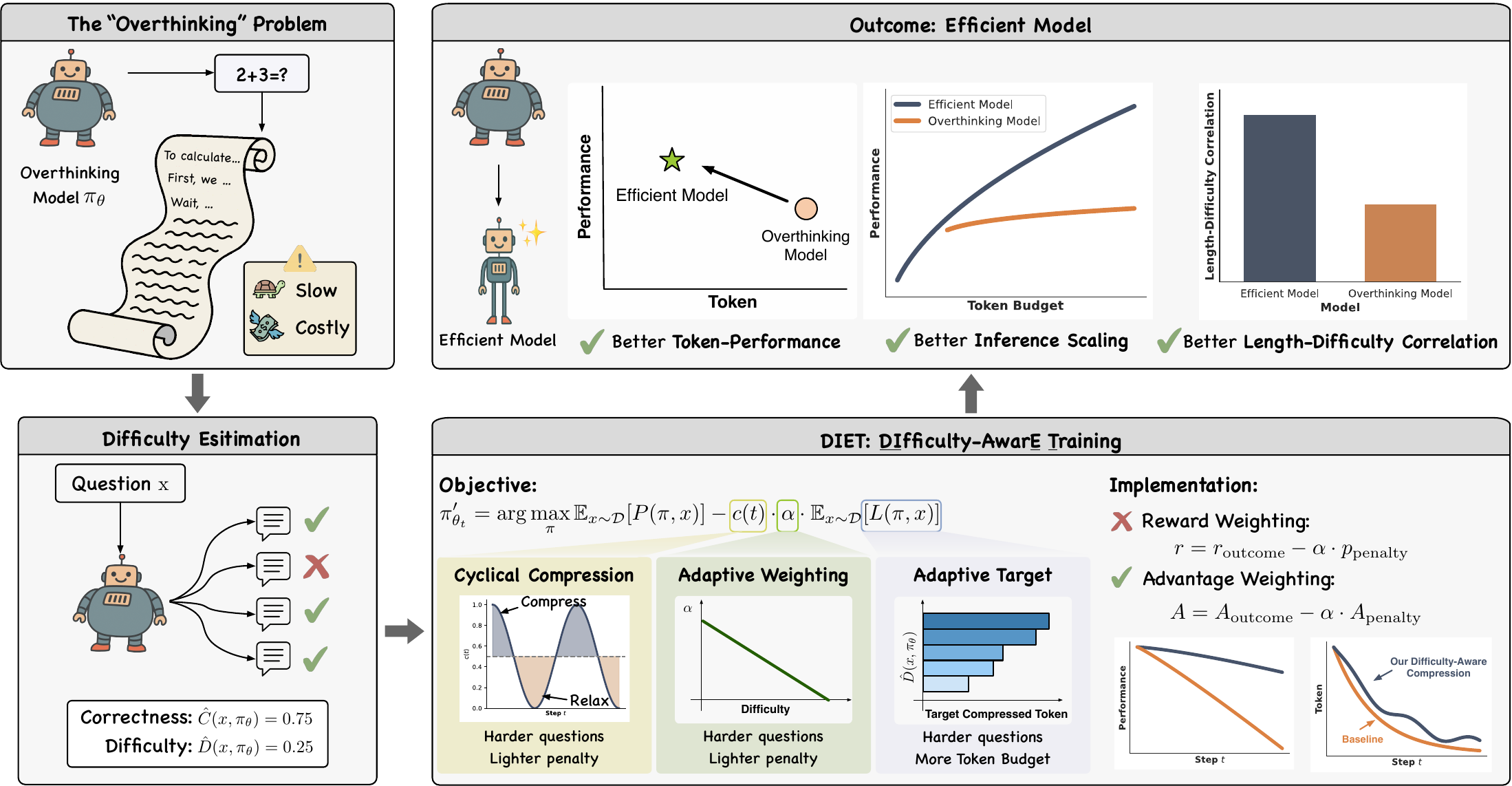}
    \vspace{-0.5em}
    \caption{An overview of \framework by mitigating LLM verbosity using difficulty-aware training.}
    \label{fig:overall}
    \vspace{-1em}
\end{figure}

\section{Problem Formulation and Preliminaries}
\label{sec:preliminary}

\subsection{Token Efficiency in LLMs with Reasoning Capabilities}
\label{sec:preliminary-efficiency}
Formally, we define the token efficiency problem studied in this paper as a multi-objective optimization challenge. Given a capable reasoning LLM policy $\pi_\theta$ and a distribution of reasoning problems $\mathcal{D}$, we aim to find a policy $\pi_\theta'$ that optimizes both performance and token efficiency:
\begin{equation}
\pi_\theta' = \arg\max_{\pi} \mathbb{E}_{x \sim \mathcal{D}} [P(\pi, x)] - \alpha \cdot \mathbb{E}_{x \sim \mathcal{D}} [f(L(\pi, x))],
\label{eq:rl-optim}
\end{equation}
where $L(\pi, x)$ represents the expected response length (in tokens) for prompt $x$ under policy $\pi$, $P(\pi, x)$ represents the performance (e.g., accuracy on math problems) on task $x$, $\alpha > 0$ is a coefficient that controls the trade-off between performance and token efficiency, and $f$ is a monotonically increasing transformation function with different implementations in prior work.

The current paradigm of RL from outcome rewards typically focuses solely on maximizing performance without considering token efficiency ($\alpha=0$), which leads to \textit{overthinking}. Previous approaches to addressing this issue have generally attempted to incorporate token length into the reward function in an intuitive manner:
\begin{equation}
r(x, y) = r_{\text{outcome}}(x, y) - \alpha \cdot f\left(L(y)\right),
\label{eq:rl-reward}
\end{equation}
where $r_{\text{outcome}}(x, y)$ represents the outcome reward, $L(y)$ is the length of response $y$ in tokens. However, the uniform penalty $\alpha$ treats all problems equally, failing to account for varying difficulty levels, and with inappropriate $\alpha$, it quickly leads to suboptimal tradeoffs, where performance degrades substantially as response length is reduced.

\subsection{Model-Based Difficulty Estimation}
\label{sec:preliminary-difficulty}
A critical insight of our work is that the optimal response length for a task should vary according to its difficulty. Intuitively, challenging problems may benefit from extended reasoning, while simpler questions can be answered concisely without sacrificing accuracy. We estimate the difficulty of a given problem based on the performance of the policy model itself during training. Formally, we define the \textit{estimated correctness} for a given prompt $x$ under policy $\pi_\theta$ based on $N$ sampled responses $\{y_i\}_{i=1}^N$:
\begin{equation}
\hat{C}(x, \pi_\theta) = \frac{1}{N} \sum_{i=1}^{N} \mathbb{I}(y_i \text{ is correct}) \quad \text{where } y_i \sim \pi_\theta(\cdot|x).
\label{eq:correctness}
\end{equation}
The estimated difficulty $\hat{D}(x, \pi_\theta)$ can then be defined as $1 - \hat{C}(x, \pi_\theta)$. This formulation captures an intuitive notion: tasks that the current policy consistently fails on ($\hat{C} \approx 0$) are considered difficult, while those consistently answered correctly ($\hat{C} \approx 1$) are considered easy.

Notably, popular RL algorithms such as GRPO~\citep{DBLP:journals/corr/abs-2402-03300} and RLOO~\citep{DBLP:conf/acl/AhmadianCGFKPUH24} already require sampling multiple responses ($N>1$) per prompt within each training batch to estimate advantages. Therefore, computing $\hat{C}(x, \pi_\theta)$ or $\hat{D}(x, \pi_\theta)$ incurs \textit{no computational overhead}, as the necessary samples and correctness evaluations are already part of the core RL algorithm. This makes on-the-fly difficulty estimation highly practical for integration into the training process, as we explore in subsequent sections.

\subsection{Preliminary Analysis: Intrinsic Correlation of Response Length and Problem Difficulty}
\label{sec:preliminary-corr-analysis}

\begin{wrapfigure}{r}{0.5\textwidth}
    \vspace{-2em}
    \begin{center}
    \includegraphics[width=\linewidth]{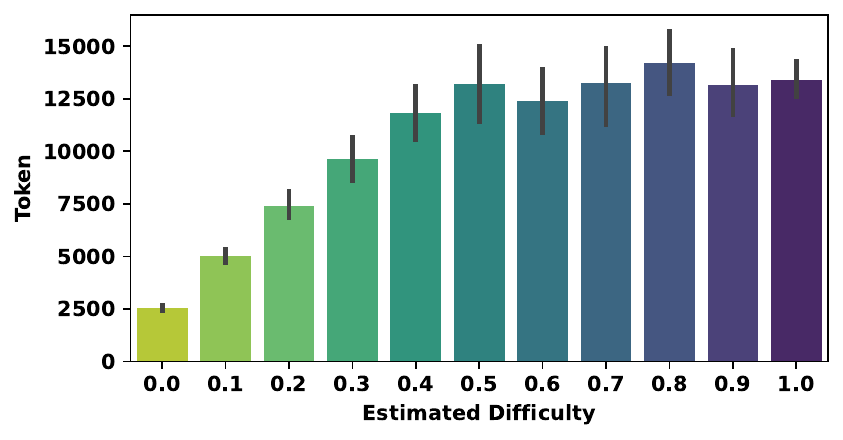}
    \end{center}
    \vspace{-1.5em}
    \caption{LLM's response length relative to problem difficulty.}
    \label{fig:base-corr}
    \vspace{-1.5em}
\end{wrapfigure}
Before introducing our compression methods, we analyze the LLM's natural verbosity relative to task difficulty. Using R1-Distill-Qwen-1.5B and difficulty estimated via \cref{eq:correctness}, we find that solution length increases with complexity, even without difficulty-aware training.

\cref{fig:base-corr} shows a strong positive correlation: average response length increases with estimated problem difficulty, similar findings have been observed in \citet{estermann2025reasoning,wu2025more}. This suggests LLMs inherently allocate more tokens to harder problems, likely for more elaborate reasoning. This pivotal observation implies that common "one-size-fits-all" compression strategies, which ignore difficulty, risk either truncating vital reasoning on complex tasks or under-compressing simple ones. Indeed, many prior compression techniques overlook this, and as shown in \cref{sec:exp_correlation_analysis}, can distort this natural length-complexity relationship post-training.

This baseline behavior is a cornerstone of our motivation. Since LLMs instinctively use more tokens for harder problems, effective compression should leverage this. We hypothesize that integrating on-the-fly difficulty awareness into compression training enables models to compress more intelligently: compress aggressively on easier problems while preserving token budgets for complex ones, which is the key to achieving superior performance-efficiency.
\section{Difficulty-Aware Reinforcement Learning for Token Compression}
\label{sec:reward-shaping}

The baseline analysis in \cref{sec:preliminary-corr-analysis} shows that LLMs naturally adjust response length based on problem difficulty. This motivates our approach: explicitly incorporating on-the-fly difficulty estimation into the RL objective. Our goal is to train models that compress responses for simpler tasks while retaining reasoning for harder ones. This section presents our methods: we first formulate how to integrate difficulty into the optimization objective, then introduce a key technique for stable implementation within policy gradients, and finally propose a strategy to refine training dynamics. An overview is shown in \cref{fig:overall}.

\subsection{Formulating Difficulty-Aware Optimization Objectives}
\label{sec:formulating_adaptive_objectives}

The standard RL objective for token compression (\cref{eq:rl-optim}) often focuses on task performance $P(\pi, x)$ with a simple, uniform trade-off $\alpha$ for response length $L(\pi, x)$. To instill difficulty awareness, we propose modifying this objective so that the pressure to be concise adapts to the estimated difficulty of prompt $x$ under the current policy $\pi_\theta$ \cref{eq:correctness}. We explore two primary strategies for this:

\subsubsection{Adaptive Trade-off Parameter \texorpdfstring{$\alpha_{\text{ada}}(x, \pi_\theta)$}{alpha\_ada(x, pi\_theta)}}
\label{sec:adaptive_alpha}

The most direct way to instill difficulty awareness into the objective from \cref{eq:rl-optim} is to make its trade-off coefficient $\alpha$ adaptive to the estimated problem difficulty. The intuition is to apply stronger pressure for conciseness (a larger $\alpha$) when the model finds the problem easy (high $\hat{C}$) and relax it when the model struggles (low $\hat{C}$).

We replace the constant $\alpha$ with an \textit{adaptive} trade-off parameter, $\alpha_{\text{ada}}(x, \pi_\theta)$, calculated using the correctness estimate $\hat{C}(x, \pi_\theta)$:
\begin{equation}
\alpha_{\text{ada}}(x, \pi_\theta) = \alpha_{\text{base}} \cdot w(\hat{C}(x, \pi_\theta)),
\label{eq:effective_alpha}
\end{equation}
where $\alpha_{\text{base}} > 0$ is a hyperparameter controlling the overall maximum strength of the efficiency objective, and $w(\cdot)$ is a monotonically increasing function mapping correctness $\hat{C} \in [0, 1]$ to a non-negative weight. A simple and effective choice is the identity function, $w(\hat{C}) = \hat{C}$, resulting in the efficiency pressure scaling linearly from $0$ for the hardest problems ($\hat{C}=0$) up to $\alpha_{\text{base}}$ for the easiest ones ($\hat{C}=1$). For the penalty function $f(L(y_i))$ itself, we adopt the formulation from \citet{team2025kimi} and denote it as $f_\text{Kimi}$:
\begin{equation}
f_\text{Kimi}(L(y_i)) =
\begin{cases}
    \gamma_i, & \text{if } \mathbb{I}(y_i\text{ is correct}) = 1 \\
    \min(0, \gamma_i), & \text{if } \mathbb{I}(y_i\text{ is correct}) = 0,
\end{cases}
\label{eq:kimi_penalty}
\end{equation}
where $\gamma_i = 0.5 - \frac{L(y_i) - \min_{j} L(y_j)}{\max_{j} L(y_j) - \min_{j} L(y_j) + \epsilon}$. The overall optimization objective effectively becomes maximizing $\mathbb{E}_{x \sim \mathcal{D}} [P(\pi, x) - \mathbb{E}_{y \sim \pi(\cdot|x)}[\alpha_{\text{ada}}(x, \pi) \cdot f_{\text{Kimi}}(L(y))]]$. This approach directly modifies the penalty strength $\alpha$ based on difficulty, prioritizing performance for difficult problems and conciseness for easy ones, while using a fixed form for $f$.

\subsubsection{Dynamic Length Target \texorpdfstring{$f_{\text{dyn}}(x, \pi_\theta)$}{f\_dyn(x, pi\_theta)}}
\label{sec:dynamic_length_target}

An orthogonal strategy is to make the penalty function $f(L(y))$ itself in \cref{eq:rl-reward} inherently adaptive to problem difficulty. Instead of relying on a fixed functional form for $f$ that only considers $L(y)$ (or $L(y_i)$ relative to peers), we now define $f$ to be explicitly conditioned on a dynamic \textit{target length}, $t(x, \pi_\theta)$. This target length depends on the estimated problem difficulty, allowing a larger verbosity "budget" for harder problems.

First, we define how the target length $t$ is determined. Based on the estimated difficulty $\hat{D}(x, \pi_\theta)$, we can sample a target $t$ from a distribution that assigns higher values for higher difficulty:
\begin{equation}
t(x, \pi_\theta) \sim \text{Uniform}(\max(0, L_{\max} \cdot (\hat{D}(x, \pi_\theta) - \delta)), L_{\max} \cdot \hat{D}(x, \pi_\theta)),
\label{eq:dynamic_target_sampling_objective}
\end{equation}
where $L_{\max}$ and $\delta$ are hyperparameters defining the maximum potential target length scale, and a buffer range (e.g., 0.1). This sampling procedure assigns shorter targets for harder problems (high $\hat{D}$) and longer targets for easier ones (low $\hat{D}$).

Next, we define our difficulty-adaptive penalty function, $f_{\text{dyn}}(y_i, x, \pi_\theta, \{y_j\})$. This function quantifies the extent to which the generated length $L(y_i)$ for response $y_i \sim \pi(\cdot|x)$ exceeds its specific target $t(x, \pi_\theta)$, normalized across $N$ sampled responses $\{y_j\}_{j=1}^N$ for the same prompt $x$. Let $p(y_i, t) = \max(0, L(y_i) - t(x, \pi_\theta))$ be the raw exceedance. Our adaptive penalty function is:
\begin{equation}
f_{\text{dyn}}(y_i, x, \pi_\theta, \{y_j\}) = \frac{p(y_i, t) - \mu_{p}}{\sigma_{p} + \epsilon},
\label{eq:normalized_penalty_objective}
\end{equation}
where $\mu_p$ and $\sigma_p$ are the mean and standard deviation of $\{p(y_j, t)\}_{j=1}^N$.
The overall optimization objective thus becomes maximizing
$\mathbb{E}_{x \sim \mathcal{D}, \{y_j\} \sim \pi(\cdot|x)} [P(\pi, x) - \alpha \cdot f_{\text{dyn}}(y_i, t(x,\pi_\theta), \{y_j\})]$.
This approach shifts the efficiency goal from minimizing absolute length to minimizing length relative to a difficulty-aware budget. These two strategies, adaptive $\alpha$ and adaptive length targets, can be explored as alternatives or potentially combined.

\subsection{Implementing Weighted Objectives with Policy Gradients: The Advantage Weighting}
\label{sec:implementation_pg}
Following the conceptual formulation of difficulty-aware objectives (\cref{sec:formulating_adaptive_objectives}), we address their RL implementation. We focus on policy gradient (PG) methods, particularly GRPO~\citep{DBLP:journals/corr/abs-2402-03300}, which uses per-prompt, multi-sample ($N>1$) advantage normalization for stability. Integrating our weighted difficulty-aware penalties requires careful analysis of weight-normalization interactions.

We show that naively combining task rewards ($r_{\text{outcome}}$) with weighted difficulty-dependent penalties into a single reward signal \textit{before} GRPO normalization causes problematic interactions. The combined reward's normalization factor (e.g., its standard deviation) then depends on both penalty and \textit{outcome variances}. Since outcome variance often correlates with problem difficulty (being highest for medium-difficulty problems and lowest for very easy or very hard ones), the intended effect of the penalty weight in the final normalized advantage becomes distorted. Specifically, as derived in \cref{app:advantage_weighting_derivation}, the token penalty is \textit{unexpectedly weakened} for problems of modest difficulty where outcome variance is high. Conversely, it can be \textit{unexpectedly exacerbated} for problems where outcome variance is very low (i.e., those the model consistently gets right or wrong), undermining the desired difficulty-aware adaptation.
\begin{remark}[Pitfall of Naive Reward Weighting]
\label{rem:naive_reward_weighting_pitfall}
In group-normalized PG algorithms like GRPO, combining reward components \textbf{before} normalization causes the penalty weight's effect to be distorted by the task's outcome variance, undermining the intended difficulty-aware adaptation.
\end{remark}
To prevent this distortion and ensure correct weighting, we propose \textit{Advantage Weighting}. This method normalizes advantages for the outcome reward ($r_{\text{outcome}, i}$) and raw penalty magnitude ($p_i$) separately, before combining them using the difficulty-aware penalty weight.
Specifically, advantages for task outcome and the penalty term are normalized independently using their per-prompt means ($\mu_{\text{outcome}}, \mu_p$) and standard deviations ($\sigma_{\text{outcome}}, \sigma_p$) from $N$ rollouts:
\begin{align}
\hat{A}_{\text{outcome}, i} = \frac{r_{\text{outcome}, i} - \mu_{\text{outcome}}}{\sigma_{\text{outcome}} + \epsilon}, \quad
\hat{A}_{p, i} = \frac{p_i - \mu_p}{\sigma_p + \epsilon}. \label{eq:normalized_component_advantages_revised}
\end{align}
Here, $p_i$ is the raw token penalty magnitude (e.g., \cref{eq:kimi_penalty,eq:normalized_penalty_objective}).
The final policy gradient advantage $\hat{A}'_i$ combines the normalized components weighted by $\alpha'$ (e.g., $\alpha_{\text{ada}}$ or $\alpha$ from \cref{sec:formulating_adaptive_objectives}):
\begin{equation}
\hat{A}'_{i} = \hat{A}_{\text{outcome}, i} - \alpha' \cdot \hat{A}_{p, i}.
\label{eq:advantage_weighting_impl_revised_short_remark}
\end{equation}
Thus, each component is scaled by its own variance \textit{before} adaptive weighting. This allows $\alpha$ to correctly modulate the normalized penalty's influence relative to that of the normalized outcome, faithfully reflecting the intended difficulty-aware trade-off and resolving the distortion.
Experiments in \cref{sec:reward-shaping-validation} validate this improvement, showing significant benefits over naive reward weighting.

\subsection{Refining Training Dynamics: Cyclical Compression Pressure}
\label{sec:training_dynamics}

Although the difficulty-aware objectives (\cref{sec:formulating_adaptive_objectives}) implemented with Advantage Weighting (\cref{sec:implementation_pg}) provide robust adaptive compression, constant pressure may risk premature convergence on brevity or hinder exploration. To address this and potentially enhance the performance-efficiency trade-off, we explore temporally modulating the compression intensity during training.

Inspired by annealing schedules, we cyclically vary the difficulty-aware penalty strength using a time-varying cosine modulation factor
$c(t) = 0.5 \left(1 + \cos\left(\frac{2\pi t}{T}\right)\right)$,
where $T$ is the cycle period. This factor $c(t)$ smoothly oscillates between $1$ (maximum pressure) and $0$ (minimum pressure) and scales the difficulty-aware penalty component within the final advantage calculation (\cref{eq:advantage_weighting_impl_revised_short_remark}):
\begin{equation}
\hat{A}'_{i}(t) = \hat{A}_{\text{outcome}, i} - c(t) \cdot \alpha_{\text{ada}}(x, \pi_\theta) \cdot \hat{A}_{p, i}.
\label{eq:cyclical_advantage_sum}
\end{equation}
This temporal variation aims to improve robustness and the final performance-efficiency trade-off, potentially allowing the model to escape local optima related to excessive brevity and better balance reasoning consolidation with conciseness pressure.

\subsection{\texorpdfstring{The \framework: \underline{DI}fficulty-Awar\underline{E} \underline{T}raining Method}{The \protect\framework: DIfficulty-AwarE Training Method}}
Our approach, \framework, effectively integrates all the elements discussed in this section. Specifically, \framework employs an objective that synergizes the principles of adaptive penalty strength (\cref{sec:adaptive_alpha}) and dynamic length target (\cref{sec:dynamic_length_target}). This combined objective is implemented using Advantage Weighting (\cref{sec:implementation_pg}) and its training is optimized with Cyclical Compression Pressure (\cref{sec:training_dynamics}). This holistic strategy, \framework, underpins the best performance-efficiency results presented in our later experiments.

\section{Experimental Validation}
\label{sec:experiments}


\subsection{Experimental Setup}
\label{sec:exp_setup}

\textbf{Base Model and Algorithm.} Our experiments use the R1-Distilled Qwen 1.5B model~\citep{DBLP:journals/corr/abs-2501-12948}. For RL algorithm, we employ GRPO~\citep{DBLP:journals/corr/abs-2402-03300}. All proposed RL methods are built on GRPO and use the Advantage Weighting technique (\cref{sec:implementation_pg}).

\textbf{Training.} We use the DeepScaleR dataset~\citep{deepscaler2025}, featuring high-quality mathematical problems of diverse complexities. We use veRL~\citep{sheng2024hybridflow} as the training framework, and train models on 8 A100 GPUs. For more details of training, please refer to \cref{app:training_details}.

\textbf{Baselines.} We compare against: (1) The \textbf{Base Model} without any compression. (2) \textbf{SFT- \& DPO-Based Methods}: Kimi 1.5 SFT~\citep{team2025kimi} (fine-tuning on shortest correct responses), Kimi 1.5 DPO~\citep{team2025kimi} (using shortest correct as positive DPO examples), and TokenSkip~\citep{DBLP:journals/corr/abs-2502-12067} (SFT on responses with redundant tokens removed). (3) \textbf{Other RL-Based Methods}: CosFn~\citep{DBLP:journals/corr/abs-2502-03373}, O1-Pruner~\citep{DBLP:journals/corr/abs-2501-12570}, and Kimi 1.5 RL~\citep{team2025kimi}.

\textbf{Evaluation.} We assess Pass@1 (P@1) and average response length (Tokens, Tok) on MATH 500~\citep{DBLP:conf/nips/HendrycksBKABTS21}, AIME 2024, AMC 2023, Olympiad Bench~\citep{DBLP:conf/acl/HeLBHTSHHHZLQL024}, and Minerva~\citep{DBLP:conf/nips/LewkowyczADDMRS22}. We sample 32 samples for each question in AIME24, and 10 samples for others to estimate the P@1. Details for evaluation are presented in \cref{app:eval_details}.

\subsection{\framework Achieves Improved Performance with Reduced Tokens}
\label{sec:exp_main_results}

\cref{tab:reward-shaping-results-1.5b} summarizes the performance and token efficiency of our methods against baselines. The Base Model establishes a strong performance benchmark but with substantial verbosity, underscoring the \textit{overthinking} issue.

\begin{table}[t]
\centering
\caption{Average performance (Pass@1, \%) and token length of R1-Distilled Qwen 1.5B trained with different token compression methods. For each benchmark, highest P@1 is bolded. For the "Macro Average" columns, bolding indicates the best P@1 and a favorable performance-efficiency trade-off.}
\label{tab:reward-shaping-results-1.5b}
\resizebox{1.0\textwidth}{!}{%
\begin{tabular}{l c c c c c c c c c c | c c}
\toprule
\multirow{2}{*}{\raisebox{-0.6\height}{\textbf{Method}}} & \multicolumn{2}{c}{\textbf{MATH 500}} & \multicolumn{2}{c}{\textbf{AIME 2024}} & \multicolumn{2}{c}{\textbf{AMC 2023}} & \multicolumn{2}{c}{\textbf{Olympiad.}} & \multicolumn{2}{c|}{\textbf{Minerva}} & \multicolumn{2}{c}{\textbf{Macro Average}} \\
\cmidrule(lr){2-3}\cmidrule(lr){4-5}\cmidrule(lr){6-7}\cmidrule(lr){8-9}\cmidrule(lr){10-11}\cmidrule(lr){12-13}
 & P@1 & Tok & P@1 & Tok & P@1 & Tok & P@1 & Tok & P@1 & Tok & P@1 & Tok \\ \midrule
Base Model & 82.1 & 5534 & 28.5 & 16590 & 62.7 & 10615 & 43.5 & 11587 & 26.0 & 7076 & 48.6 & 10280 \\ \midrule
\textbf{\textit{SFT- \& DPO-Based}}\\
Kimi 1.5 SFT & 68.5 & 6761 & 22.0 & 17400 & 60.4 & 9323 & 39.4 & 10036 & 23.6 & 2804 & 42.7$_{\textcolor{myred}{-12.1\%}}$ & 9865$_{\ \textcolor{mygreen}{-4.0\%}}$ \\
Kimi 1.5 DPO & \textbf{83.3} & 4464 & 31.7 & 13389 & 63.0 & 8678 & \textbf{44.5} & 9604 & 26.9 & 6070 & 49.9$_{\ \textcolor{mygreen}{+2.7\%}}$ & 8441$_{\textcolor{mygreen}{-17.9\%}}$ \\
TokenSkip & 64.1 & 1120 & 6.8 & 2231 & 37.3 & 1401 & 25.8 & 2061 & 20.7 & 1674 & 30.9$_{\textcolor{myred}{-36.4\%}}$ & 1697$_{\textcolor{mygreen}{-83.5\%}}$ \\
\midrule
\textbf{\textit{RL-Based}}\\
CosFn & 75.6 & 2735 & 27.5 & 12492 & 61.1 & 6970 & 42.9 & 8307 & \textbf{27.1} & 3485 & 46.8$_{\textcolor{myred}{-3.5\%}}$ & 6798$_{\textcolor{mygreen}{-33.9\%}}$ \\
O1-Pruner & 79.1 & 2531 & 25.0 & 8961 & 62.5 & 5010 & 39.0 & 5242 & 23.7 & 2400 & 45.9$_{\textcolor{myred}{-5.4\%}}$ & 4829$_{\textcolor{mygreen}{-53.0\%}}$ \\
Kimi 1.5 RL & 66.3 & 1552 & 18.8 & 9109 & 44.7 & 3808 & 28.5 & 4774 & 16.7 & 1009 & 35.0$_{\textcolor{myred}{-27.9\%}}$ & 4050$_{\textcolor{mygreen}{-60.6\%}}$ \\
\midrule
\textbf{\textit{Our Difficulty-Aware Methods}}\\
Dynamic Target (\cref{sec:dynamic_length_target}) & 82.1 & 2792 & 27.7 & 10288 & 63.4 & 6017 & 43.4 & 6490 & 26.3 & 2700 & 48.6$_{\textcolor{myyellow}{+0.0\%}}$ & 5657$_{\textcolor{mygreen}{-45.0\%}}$ \\
Adaptive Weighting (\cref{sec:adaptive_alpha}) & 82.7 & 2876 & \textbf{32.2} & 10255 & 64.4 & 5819 & 43.7 & 6494 & 26.6 & 3170 & 49.9$_{\textcolor{mygreen}{+2.8\%}}$ & 5723$_{\textcolor{mygreen}{-44.3\%}}$ \\
DIET (\cref{sec:adaptive_alpha}+\cref{sec:dynamic_length_target}) & 83.0 & 3061 & 31.8 & 10578 & \textbf{65.4} & 6425 & 43.7 & 6917 & 26.9 & 3505 & \textbf{50.2}$_{\textcolor{mygreen}{+3.3\%}}$ & \textbf{6097}$_{\textcolor{mygreen}{-40.7\%}}$ \\
\bottomrule
\end{tabular}%
}
\vspace{-1.9em}
\end{table}

Among existing approaches, Kimi 1.5 DPO is a strong baseline that slightly reduces tokens and improves the performance over the base model. TokenSkip achieves extreme token reduction, but its average P@1 drops significantly, demonstrating that aggressive, non-nuanced compression severely degrades reasoning. Standard RL baselines generally achieve more significant token reduction, however, the performances are slightly inferior to the base model. These methods highlight the difficulty of achieving both high performance and low token counts simultaneously without more sophisticated adaptation.

Our \textbf{Difficulty-Aware Methods} consistently yield a better performance-efficiency frontier. The \textit{Dynamic Target} approach maintains the Base Model's Macro Average P@1 while reducing average tokens by a substantial 45.0\%. The \textit{Adaptive Weighting} method further improves the Macro Average P@1 to 49.9\% ($+2.8\%$ over Base), matching Kimi 1.5 DPO's performance but with considerably fewer tokens. Significantly, \framework, which leverages both Dynamic Target and Adaptive Weighting, achieves the best P@1. This performance is achieved with an average token count of only 6097, a significant 40.7\% reduction compared to the Base Model. Beyond these quantitative gains, our qualitative analysis (\cref{app:case_study}) further reveals that \framework training progressively refines the model's reasoning style, leading to more structured language, concise calculations, and a marked reduction in unnecessary self-doubt and redundant post-solution exploration.

These results underscore the effectiveness of incorporating nuanced difficulty awareness. While methods like TokenSkip can produce very short answers, they do so at an unacceptable performance drop. Our difficulty-aware methods, especially the combined strategy, demonstrate that it is possible to achieve substantial token reductions while maintaining, and even enhancing the sophisticated reasoning capabilities of the LLM, leading to a superior performance-efficiency trade-off.

\subsection{\framework's Advantage in Inference Scaling}
\label{sec:exp_inference_scaling}

\begin{wrapfigure}{r}{0.45\textwidth}
    \vspace{-2em}
    \begin{center}
      \includegraphics[width=1.0\linewidth]{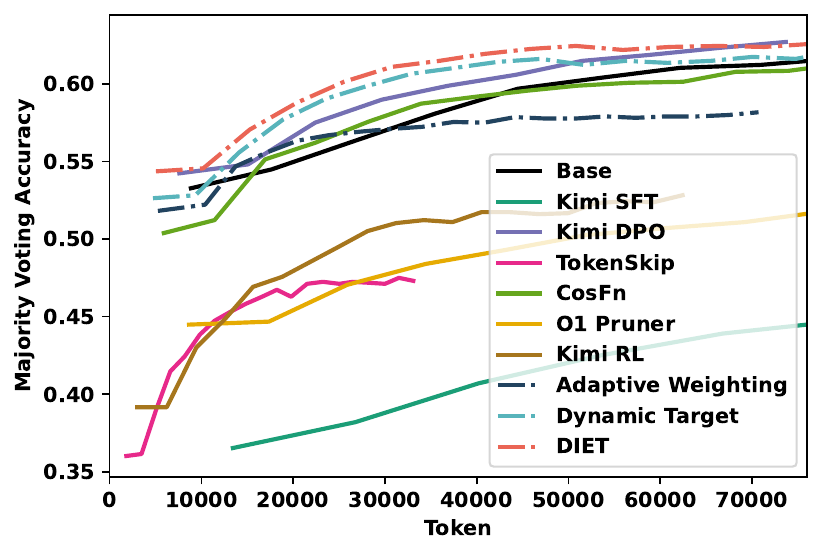}
    \end{center}
    \vspace{-1.3em}
    \caption{Micro average of majority voting Pass@1 on all the benchmarks.}
    \label{fig:inference_scaling_aime}
    \vspace{-1.5em}
\end{wrapfigure}

An often-overlooked benefit of token compression is its potential to enhance inference scaling performance under a fixed total token budget. Shorter responses allow for more samples to be drawn for techniques like majority voting, which can improve overall accuracy if per-sample quality is maintained. While previous token compression work typically focused on single-sample token counts and Pass@1, we investigate this inference scaling behavior. We show that many existing compression methods fail to translate token savings into improved scaling performance, whereas \framework achieve superior majority voting accuracy, particularly at practical, lower token budgets.

\cref{fig:inference_scaling_aime} plots the micro average of majority voting Pass@1 across all math benchmarks against the total token budget. The results highlight the advantages of our approaches. Methods like TokenSkip and Kimi SFT, despite allowing many samples due to extreme compression, exhibit low and quickly stagnating majority voting accuracy, confirming that their severe per-sample quality degradation undermines scaling benefits. While stronger baselines like Kimi DPO show more respectable scaling, their increase in the scaling performance is slow. The Base Model itself requires a substantial token budget before multiple samples yield significant gains.

In contrast, our difficulty-aware methods demonstrate advantages. Notably, \framework and Dynamic Target achieve significantly higher majority voting accuracy at low token budgets. This demonstrates that \framework effectively preserves per-sample quality, allowing the benefits of increased sampling to manifest early. Although the final accuracies are similar to other baselines, the faster convergence makes its scaling more practical.
Overall, the results show that our methods effectively translate token savings into improved inference scaling performance, which is crucial for practical applications.

\subsection{\framework Enhances Length-Difficulty Correlation}
\label{sec:exp_correlation_analysis}

An ideal token compression method should not only reduce verbosity but also preserve, or even enhance, the intelligent allocation of tokens based on problem difficulty, a natural tendency observed in base LLMs (\cref{sec:preliminary-corr-analysis}). Uniform compression risks disrupting this by being overly terse on complex problems or insufficiently concise on simple ones. We investigate this by measuring the Pearson correlation between estimated problem difficulty and generated response length across our test benchmarks; a higher positive correlation indicates more appropriately scaled verbosity.

\begin{wrapfigure}{r}{0.4\textwidth}
    \vspace{-1.8em}
    \begin{center}
        \includegraphics[width=1.0\linewidth]{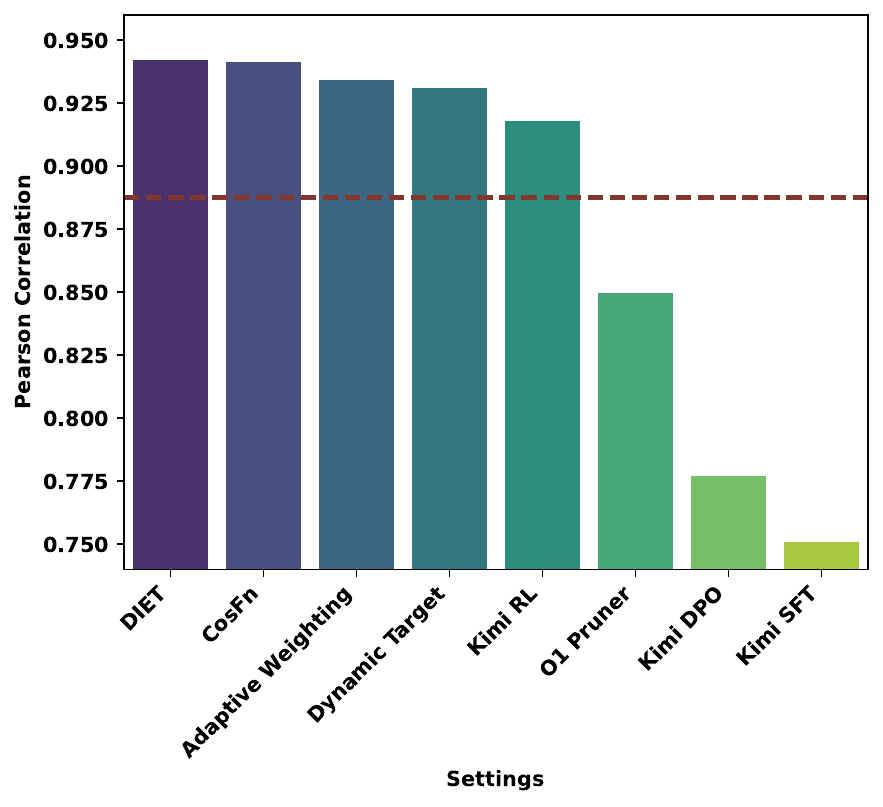}
    \end{center}
    \vspace{-1em}
    \caption{Pearson correlation between problem difficulty and average response length of different methods. All p-values are lower than 0.01. Methods that are not included have lower correlation.}
    \label{fig:correlation}
    \vspace{-1em}
\end{wrapfigure}
\cref{fig:correlation} demonstrates that our difficulty-aware methods excel at maintaining and enhancing this crucial correlation. Our method, \framework, achieves the highest correlation, surpassing the Base Model's inherent correlation (dashed horizontal line) and other baselines. This indicates that \framework successfully learns to modulate verbosity in tight alignment with problem difficulty, consistent with our design goals. Notably, another strong baseline, CosFn, also shows a high correlation, nearly matching \framework, but as shown in \cref{sec:exp_main_results}, it fails to preserve performance when reducing tokens. In contrast, other compression techniques significantly degrade this adaptive characteristic. For instance, Kimi DPO and Kimi SFT show a markedly weaker correlation than the Base Model, implying their compression mechanisms are less sensitive to problem difficulty. O1 Pruner also shows a reduced correlation. This suggests that while these methods reduce tokens, they do so in a more uniform or difficulty-agnostic manner.

These results underscore that intelligent compression, as achieved by \framework, is not merely about token reduction but about doing so in a way that respects problem complexity. By strengthening the positive relationship between difficulty and response length, \framework ensures a more rational allocation of computational budget during inference, contributing to its robust performance-efficiency.

\begin{figure}[t]
    \centering
    \begin{subfigure}[b]{0.47\textwidth}
        \centering
        \includegraphics[width=\linewidth]{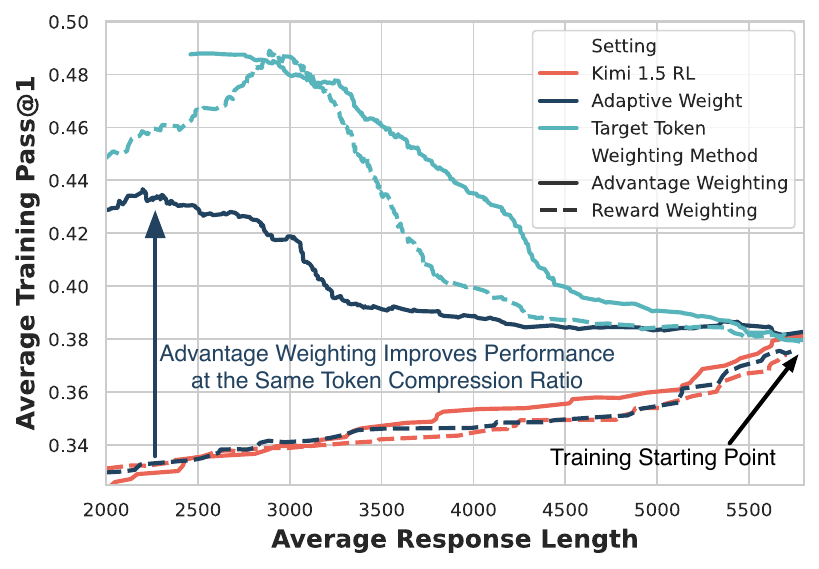}
    \end{subfigure}
    \hfill
    \begin{subfigure}[b]{0.47\textwidth}
        \centering
        \includegraphics[width=\linewidth]{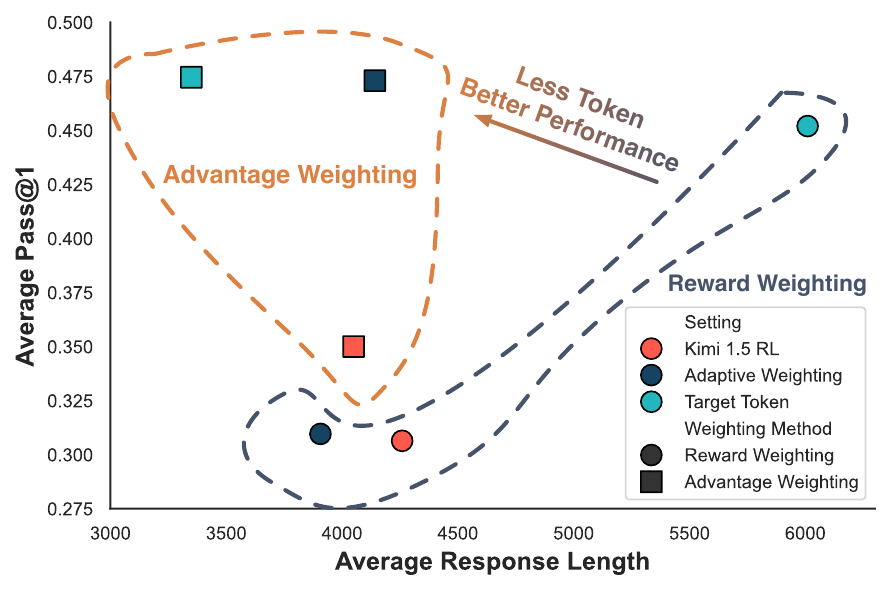}
    \end{subfigure}
    \vspace{-1em}
    \caption{\textbf{Advantage Weighting vs. Reward Weighting analysis. (Left)} Training curves (Pass@1 vs. Response Length) demonstrate better performance with Advantage Weighting. \textbf{(Right)} Final evaluation results show Advantage Weighting yields superior performance-efficiency points.}
    \label{fig:advantage_weighting_exp}
    \vspace{-1.8em}
\end{figure}

\subsection{Ensuring Stable Difficulty-Aware RL Training with Advantage Weighting}
\label{sec:reward-shaping-validation}

We empirically validate our proposed Advantage Weighting method (\cref{sec:implementation_pg}), designed to overcome the signal distortion pitfalls of naive reward weighting within normalized policy gradient algorithms like GRPO. As highlighted in our analysis (\cref{sec:implementation_pg}), naive weighting can interact poorly with outcome variance, hindering effective penalty application.

\cref{fig:advantage_weighting_exp} (left) plots the training dynamics (Average Training Pass@1 vs. Average Response Length). Advantage Weighting (solid lines) consistently maintains higher performance than Reward Weighting (dashed lines) as models compress responses (moving right-to-left during training). This demonstrates more effective learning during token reduction across all penalty settings when using Advantage Weighting.
The final evaluation results (\cref{fig:advantage_weighting_exp}, right) further reinforce this. Models trained with Advantage Weighting achieve superior performance-efficiency trade-offs, occupying the desirable top-left region (high performance, low token count). The \framework without advantage weighting fails to deliver reasonable performance, we therefore omit it from the visualization.

These empirical results strongly support our theoretical analysis (\cref{sec:implementation_pg}). Advantage Weighting is crucial for the stable and effective implementation of weighted objectives (like difficulty-aware penalties) in normalized PG algorithms. Naive reward weighting leads to suboptimal training and significantly poorer final performance-efficiency outcomes.

\section{Related Work}
\label{sec:relwork}
Efficient reasoning in LLMs has recently attracted significant attention, as methods that boost reasoning performance, often via RL with outcome-based rewards, unintentionally induce verbose, overthought outputs. A growing body of work has therefore aimed at compressing the reasoning to reduce inference costs without greatly compromising accuracy.

\paragraph{Prompt-Based Approaches.}
Initial efforts primarily explore prompt engineering to reduce verbosity. For example, Chain-of-Draft~\citep{DBLP:journals/corr/abs-2502-18600} and Sketch-of-Thought~\citep{aytes2025sketch} restructure reasoning by having the model first draft a concise outline before finalizing the answer, while Constrained-CoT~\citep{DBLP:journals/corr/abs-2407-19825} imposes length limits via prompts. Though these methods allow quick, zero-shot adjustments, their effectiveness is limited since they do not alter the model's internal parameters. 

\paragraph{Training-Based Compression via Supervised and RL Methods.}
More fundamental approaches modify the model's training process to inherently produce more concise reasoning chains. Supervised fine-tuning (SFT) techniques aim to internalize efficient reasoning patterns by training on compressed or optimized CoT data. SPIRIT-FT~\citep{DBLP:journals/corr/abs-2502-13260} and Skip-Steps~\citep{DBLP:conf/nips/LiuGHJZQZ24} train models on reasoning steps deemed crucial. Other SFT approaches focus on distilling longer reasoning chains from capable models into shorter, equivalent ones~\citep{DBLP:journals/corr/abs-2407-06023,DBLP:journals/corr/abs-2412-11664,DBLP:journals/corr/abs-2502-20122}. Some methods even train models to reason implicitly in latent space, generating concise outputs without explicit step-by-step textual reasoning, such as Coconut~\citep{DBLP:journals/corr/abs-2412-06769}, CCoT~\citep{DBLP:journals/corr/abs-2412-13171}, and Implicit-CoT~\citep{DBLP:journals/corr/abs-2405-14838}.

Reinforcement learning (RL) offers another avenue, often by incorporating length penalties directly into the reward function. This typically involves combining the primary outcome-based reward (e.g., correctness) with a secondary reward term that penalizes longer sequences. Examples include approaches by~\citet{DBLP:journals/corr/abs-2502-04463}, L1~\citep{aggarwal2025l1}, Kimi-1.5~\citep{team2025kimi}, and work exploring how length penalties can be used to stabilize training~\citep{DBLP:journals/corr/abs-2502-03373}. Our work builds on these efforts by comprehensively introducing difficulty awareness into the compression process, analyzing the impact of data difficulty on compression training, proposing adaptive reward shaping techniques that dynamically adjust token penalties based on on-the-fly difficulty estimation, and addressing methodological issues in applying such rewards within group-normalized RL methods.

\section{Conclusion}
\label{sec:conclusion}

To combat LLM "overthinking" and its inherent inefficiencies, we introduced \textbf{\framework (\underline{DI}fficulty-Awar\underline{E} \underline{T}raining)}, a framework that intelligently "cuts token calories" by integrating on-the-fly problem difficulty into the RL process for adaptive compression. \framework achieves satisfactory reasoning performance while significantly reducing token counts. Beyond these primary gains, \framework uniquely preserves and enhances the natural positive correlation between response length and problem difficulty, ensuring appropriate verbosity. Furthermore, it translates these efficiencies into superior \textbf{inference scaling}, delivering better performance under fixed computational budgets, which is a crucial advantage over prior methods that often falter in this regard. \framework thus offers a principled, effective, and thoroughly-validated strategy for developing more practical, efficient, and ultimately more capable large language models.

\bibliographystyle{reference}
\bibliography{reference}


\appendix

\section{Limitations}
\label{app:limitations}

While \framework demonstrates significant advancements in creating more token-efficient and performant reasoning LLMs, we acknowledge certain limitations.

Our empirical validation has primarily focused on mathematical reasoning benchmarks. Although these tasks robustly test complex reasoning and verbosity patterns, the generalization of \framework's benefits to other diverse domains warrants more extensive investigation. The optimal balance between conciseness and necessary verbosity might vary across these different applications. Still, we think that mathematical reasoning is a representative reasoning task that can be used to validate the effectiveness of our methods.

Furthermore, while the principles of our difficulty-aware framework are conceptually orthogonal to many existing RL-based token compression techniques, suggesting potential for synergistic combinations, this work did not investigate such hybrid approaches. The empirical exploration of combining \framework with other methods to potentially achieve further performance enhancements remains an avenue for future research.

Addressing these areas could lead to even more versatile and efficient large language models.

\section{Derivation of Advantage Distortion under Naive Reward Weighting}
\label{app:advantage_weighting_derivation}

This appendix provides the mathematical details supporting Remark~\ref{rem:naive_reward_weighting_pitfall} in \cref{sec:implementation_pg}, demonstrating how naive reward weighting before normalization in algorithms like GRPO distorts the intended effect of adaptive penalty weights.

Consider implementing an objective with an adaptive trade-off parameter $\alpha_{\text{ada}}(x, \pi_\theta)$ as defined in \cref{sec:adaptive_alpha}. The naive approach combines the outcome reward $r_{\text{outcome}, i}$ and the penalty term $p_i$ for response $y_i$ to prompt $x$ into a single reward:
\begin{equation}
r'_i = r_{\text{outcome}, i} - \alpha_{\text{ada}}(x, \pi_\theta) \cdot p_i.
\label{eq:app_naive_weighted_reward}
\end{equation}
GRPO then computes the normalized advantage based on the empirical mean $\mu_{r'}$ and standard deviation $\sigma_{r'}$ of these combined rewards $\{r'_j\}_{j=1}^N$ for the $N$ responses sampled for prompt $x$:
\begin{equation}
\hat{A}'_{i} = \frac{r'_i - \mu_{r'}}{\sigma_{r'} + \epsilon}.
\label{eq:app_grpo_advantage_combined}
\end{equation}
Let $\mu_{\text{outcome}}, \sigma_{\text{outcome}}$ and $\mu_p, \sigma_p$ be the empirical means and standard deviations of the outcome rewards and penalty terms, respectively, within the batch for prompt $x$. Then $\mu_{r'} = \mu_{\text{outcome}} - \alpha_{\text{ada}} \mu_p$. Assuming $r_{\text{outcome}}$ and $p$ are approximately independent given the prompt within the batch, the variance of the combined reward is:
\begin{equation}
\sigma_{r'}^2 = \text{Var}(r'_i) = \text{Var}(r_{\text{outcome}, i} - \alpha_{\text{ada}} p_i) \approx \text{Var}(r_{\text{outcome}, i}) + \alpha_{\text{ada}}^2 \text{Var}(p_i) = \sigma_{\text{outcome}}^2 + \alpha_{\text{ada}}^2 \sigma_p^2.
\label{eq:app_combined_variance}
\end{equation}
Substituting the mean and standard deviation into the advantage calculation:
\begin{align}
\hat{A}'_{i} &= \frac{(r_{\text{outcome}, i} - \alpha_{\text{ada}} p_i) - (\mu_{\text{outcome}} - \alpha_{\text{ada}} \mu_p)}{\sqrt{\sigma_{\text{outcome}}^2 + \alpha_{\text{ada}}^2 \sigma_p^2} + \epsilon} \nonumber \\
&= \frac{(r_{\text{outcome}, i} - \mu_{\text{outcome}}) - \alpha_{\text{ada}} (p_i - \mu_p)}{\sqrt{\sigma_{\text{outcome}}^2 + \alpha_{\text{ada}}^2 \sigma_p^2} + \epsilon} \nonumber \\
&= \underbrace{\frac{1}{\sqrt{\sigma_{\text{outcome}}^2 + \alpha_{\text{ada}}^2 \sigma_p^2 + \epsilon'}}}_{\text{outcome scaling}} (r_{\text{outcome}, i} - \mu_{\text{outcome}}) - \underbrace{\frac{\alpha_{\text{ada}}}{\sqrt{\sigma_{\text{outcome}}^2 + \alpha_{\text{ada}}^2 \sigma_p^2 + \epsilon'}}}_{\text{Effective Penalty Scaling: } \hat{\tau}_p} (p_i - \mu_p).
\label{eq:app_naive_advantage_decomp}
\end{align}
The crucial term is the effective penalty scaling factor $\hat{\tau}_p$. This factor dictates how strongly the centered penalty term $(p_i - \mu_p)$ contributes to the advantage signal and the subsequent policy gradient update $\nabla_\theta J \propto \mathbb{E}[\nabla_\theta \log \pi_\theta \cdot \hat{A}']$.

Critically, $\hat{\tau}_p$ depends on the task outcome variance $\sigma_{\text{outcome}}^2$. For binary outcome rewards (correct/incorrect), $\sigma_{\text{outcome}}^2 = \hat{C}(1-\hat{C})$, where $\hat{C}$ is the estimated correctness (Eq.~\ref{eq:correctness}). This introduces an unintended dependency on problem difficulty:

\begin{itemize}[noitemsep,topsep=0pt,parsep=0pt,partopsep=0pt,leftmargin=1.5em]
    \item \textbf{Easy/Hard Problems ($\hat{C} \approx 1$ or $0$):} In these cases, the outcome variance $\sigma_{\text{outcome}}^2 \approx 0$. The effective penalty scaling becomes $\hat{\tau}_p \approx \frac{\alpha_{\text{ada}}}{\sqrt{\alpha_{\text{ada}}^2 \sigma_p^2}} = \frac{\alpha_{\text{ada}}}{|\alpha_{\text{ada}}| \sigma_p} = \frac{1}{\sigma_p}$ (assuming $\alpha_{\text{ada}} > 0$). The intended adaptive weight $\alpha_{\text{ada}}$ cancels out relative to the penalty's own standard deviation. We aim to apply strong pressure ($\alpha_{\text{ada}} \approx \alpha_{\text{base}}$) on easy problems and weak or zero pressure on hard problems. However, the normalization removes this explicit control signal strength, scaling the penalty only by its own variance $\sigma_p$.
    \item \textbf{Intermediate Difficulty Problems ($\hat{C} \approx 0.5$):} Here, the outcome variance $\sigma_{\text{outcome}}^2$ is maximal ($\approx 0.25$ for binary rewards). The denominator $\sqrt{\sigma_{\text{outcome}}^2 + \alpha_{\text{ada}}^2 \sigma_p^2}$ is larger, meaning the effective penalty scaling $\hat{\tau}_p$ is minimized. The penalty signal's contribution to the advantage is suppressed most strongly precisely when the task difficulty is intermediate.
\end{itemize}

This analysis reveals that normalizing the combined reward $r'$ distorts the effect of the adaptive weight $\alpha_{\text{ada}}$. The interaction with the difficulty-dependent outcome variance $\sigma_{\text{outcome}}^2$ prevents $\alpha_{\text{ada}}$ from cleanly controlling the relative strength of the penalty term in the advantage estimate. The policy gradient updates do not accurately reflect the desired difficulty-aware trade-off, potentially hindering convergence.

The \textit{Advantage Weighting} approach presented in \cref{sec:implementation_pg} avoids this issue by normalizing the outcome and penalty advantages separately \textit{before} applying the weight $\alpha_{\text{ada}}$, thus preserving the intended adaptive scaling.

\section{Training Details}
\label{app:training_details}
For all baseline methods, we follow the hyperparameter settings reported in their original implementations. For our RL-based methods, in the rollout phase, we set the number of rollouts to 8, with a top-p value of 0.95, a temperature of 0.6, and a maximum response length of 8192 tokens. During the training phase, we set $\alpha_{base}$ in \cref{eq:effective_alpha} to 0.5, half-cycle of Cyclical Compression Pressure to 100, kl loss coefficient to 0.001, the learning rate to 1e-6, and the batch size to 128.

\section{Evaluation Detailes}
\label{app:eval_details}
\text This section of the appendix provides details regarding the evaluation model.

\textbf{Parameters used for Evaluation:} During the evaluation, we employed a temperature of 0.6, a top-p value of 0.95, and a maximum response length of 32,768.

\textbf{Sample Count for Different Datasets:} For the MATH 500, AMC 2023, Olympiad, and Minerva datasets, we adopted a sample count of 10; for the AIME 2024 dataset, we adopted a sample count of 32.

\textbf{Method of Calculating Pass@1, and Token:} For each question, we considered the average accuracy of every sample as Pass@1, the average response length of every sample as Token.

\textbf{Prompt Used in Evaluation:} We utilized the prompt "<Question> Let's think step by step and output the final answer within \textbackslash\textbackslash boxed\{ \}" during the evaluation.

\textbf{Inference Scaling:} During the Inference Scaling evaluation of mathematical problems, we utilized Python’s \texttt{sympy}\footnote{\url{https://www.sympy.org/}} module to ascertain the equivalence of two mathematical formulas in LaTeX format. We group the responses with equivalent answers, and select the largest group as the majority voting result. We adopted the LaTeX mathematical formula that appeared most frequently in the $k$ samples after transformation by sympy as the result of Majority Voting. The Inference Scaling Accuracy was computed based on whether the Majority Voting result was equivalent to the Ground Truth.

\section{Qualitative Analysis of Behavioral Changes During Training}
\label{app:case_study} 

To understand how our \framework training qualitatively refines the model's verbosity and reasoning style beyond aggregate token counts, we conducted a case study. We aimed to assess specific behavioral changes related to token reduction by defining four qualitative metrics:
\begin{enumerate}[noitemsep,topsep=0pt,parsep=0pt,partopsep=0pt,leftmargin=1.5em]
    \item \textbf{Reduced Unnecessary Self-Doubt:} The model exhibits less hesitation or redundant self-correction once a correct reasoning path is identified.
    \item \textbf{Reduced Post-Solution Exploration:} The model curtails exploration of alternative methods or further elaboration after a correct answer has already been found.
    \item \textbf{Improved Language Structure:} The model's output is more organized and flows logically, with fewer digressions or poorly structured sentences.
    \item \textbf{Concise Calculation Process:} Mathematical or logical steps are presented more directly and with less intermediate clutter.
\end{enumerate}
For this analysis, we use questions from AMC 2023, AIME 2024, and MATH 500, and compare responses from the base model against those from various checkpoints of our \framework model during its training process. To evaluate the relative improvement on the aforementioned characteristics, we utilized Gemini 2.5 Pro as a judge to determine if each checkpoint response demonstrated the targeted behavior when compared with response from the base model.

\begin{figure}[t]
    \centering
    \includegraphics[width=0.8\textwidth]{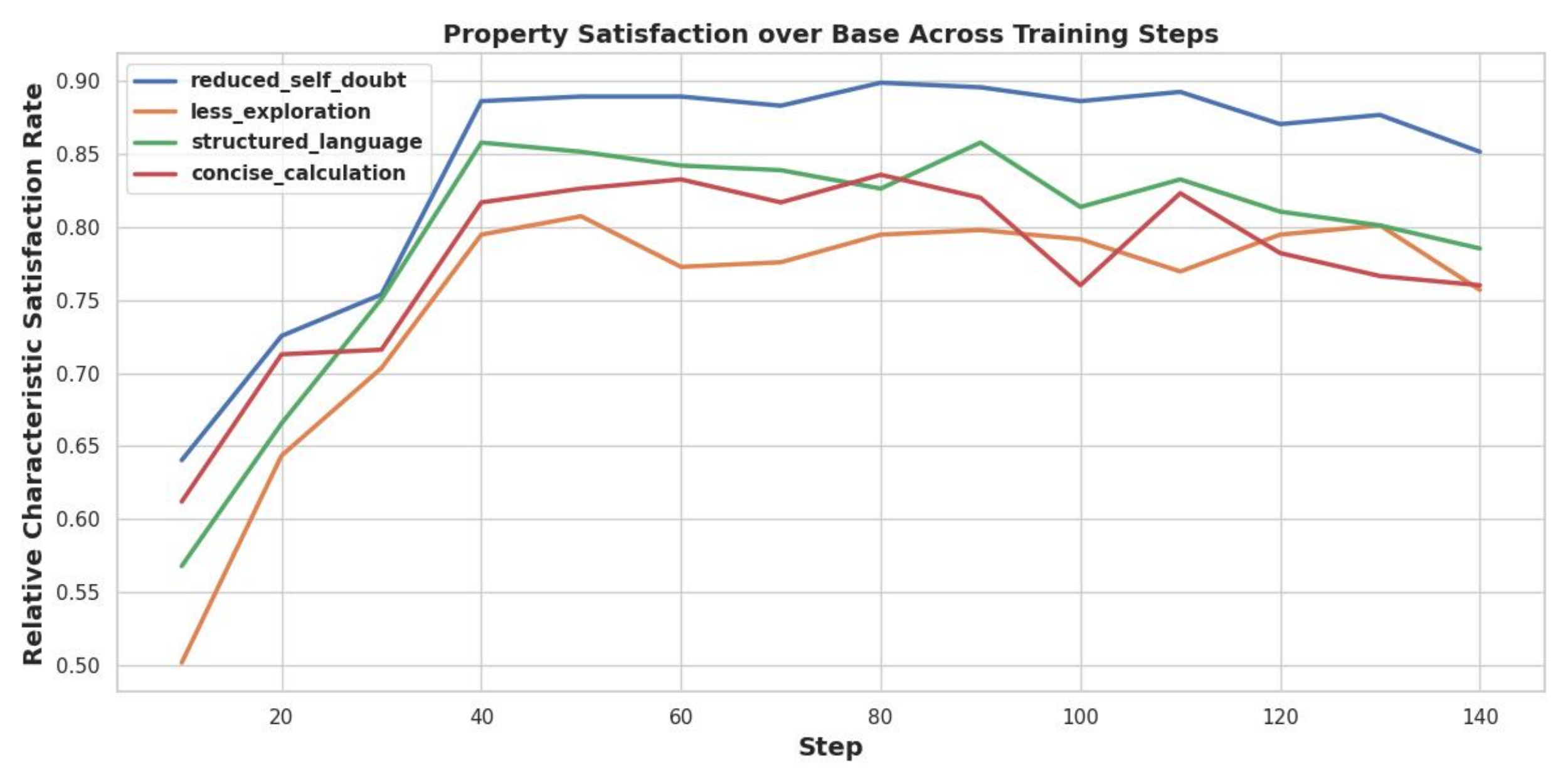}
    \caption{Evolution of qualitative characteristics during \framework training, showing the average rate at which model checkpoints satisfy each property better than the Base Model.}
    \label{fig:case_study_stats}
\end{figure}

\cref{fig:case_study_stats} illustrates the average "Property Satisfaction Rate" for these four characteristics as training progresses. Each curve represents the proportion of cases where the \framework checkpoint was judged superior to the Base Model for that specific characteristic, averaged across datasets.

The trends in \cref{fig:case_study_stats} indicate that as \framework training advances, the model progressively improves across these qualitative dimensions. We observe a clear learning curve where the model becomes more adept at producing concise and well-structured language. Notably, the "Reduced Self-Doubt" characteristic shows the most significant impact, with satisfaction rates reaching approximately 90\% by the later stages of training. "Structured Language" and "Concise Calculation" also demonstrate substantial gains, with satisfaction rates plateauing around 85\% and 83\%, respectively. "Less Exploration" after finding a solution also improves steadily, reaching around 80\%. These qualitative improvements suggest how \framework shortens the reasoning trajectory.

\section{Ablation Studies}
\label{app:ablation}
\subsection{Impact of Cyclical Compression Pressure}
\label{app:ablation_cyclical}

We analyze the effect of the cyclical compression pressure strategy, detailed in \cref{sec:training_dynamics}, by comparing our primary difficulty-aware approaches: Adaptive Weighting, Dynamic Target, and our combined \framework method with and without this temporal modulation.

\begin{figure}[htp]
  \centering
  \begin{minipage}{0.33\textwidth}
    \centering
    \includegraphics[width=\linewidth]{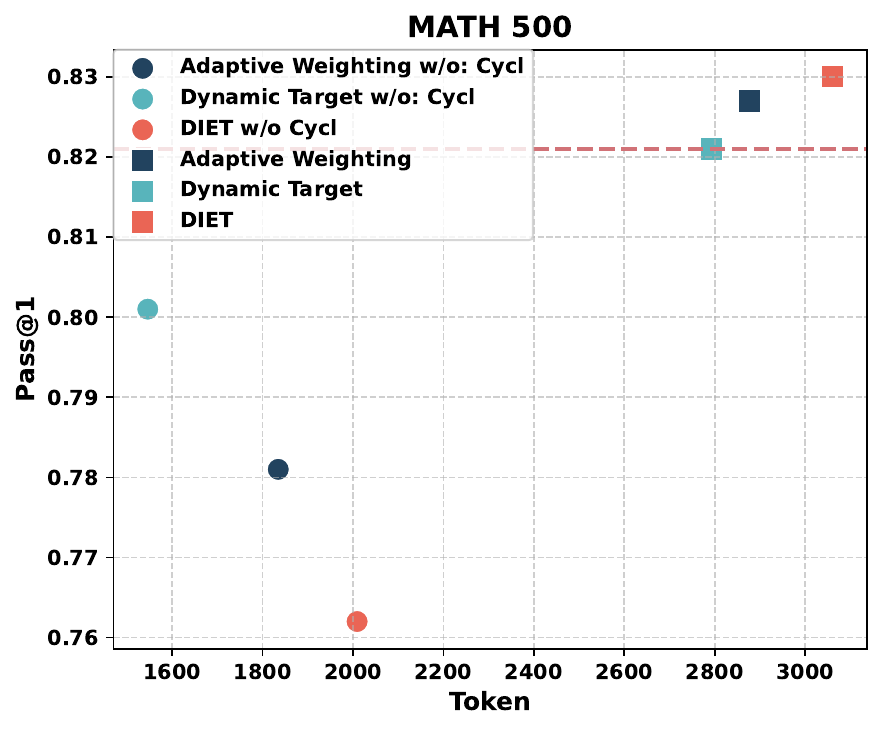}
  \end{minipage}%
  \begin{minipage}{0.33\textwidth}
    \centering
    \includegraphics[width=\linewidth]{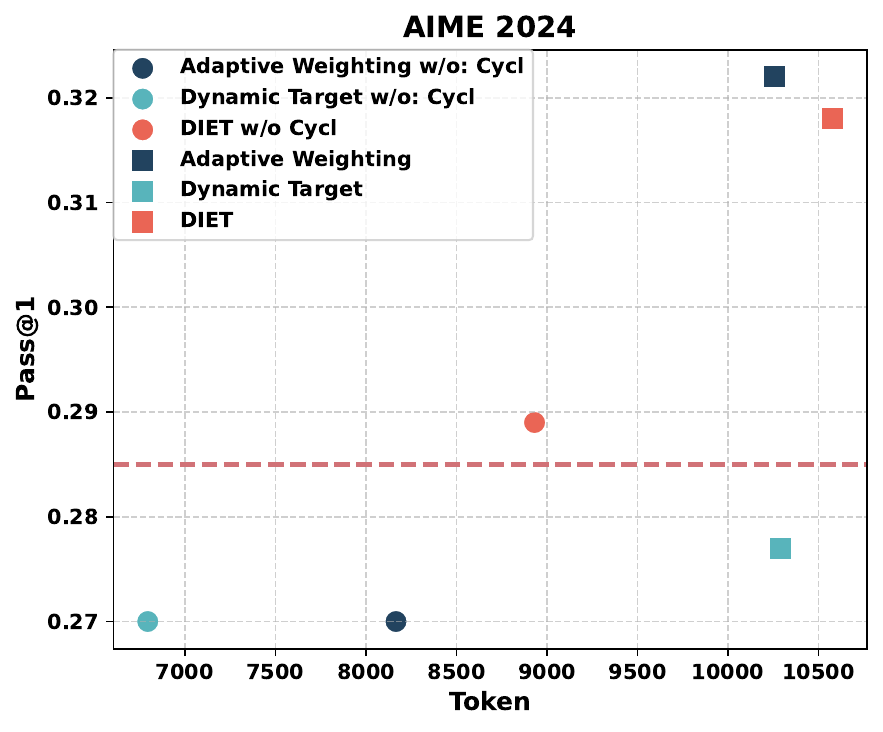}
  \end{minipage}%
  \begin{minipage}{0.33\textwidth}
    \centering
    \includegraphics[width=\linewidth]{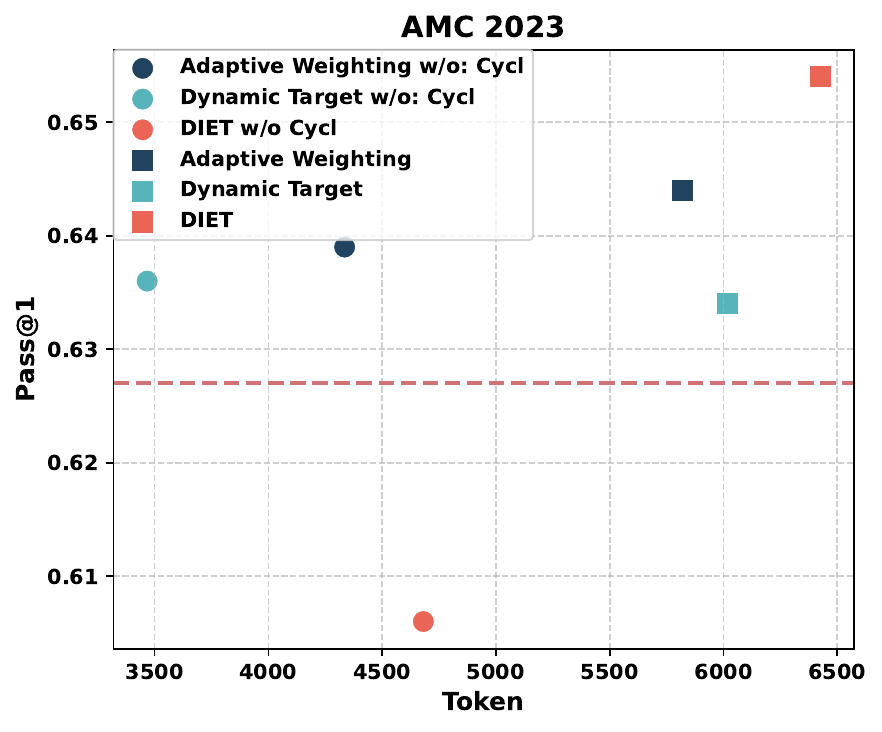}
  \end{minipage}

  \begin{minipage}{0.33\textwidth}
    \centering
    \includegraphics[width=\linewidth]{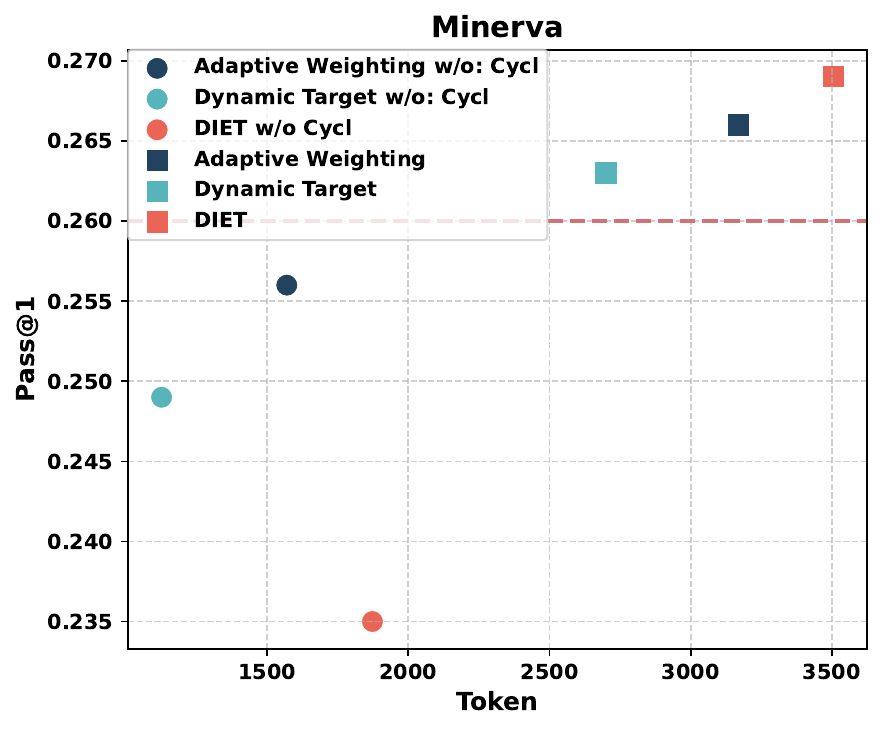}
  \end{minipage}%
  \begin{minipage}{0.33\textwidth}
    \centering
    \includegraphics[width=\linewidth]{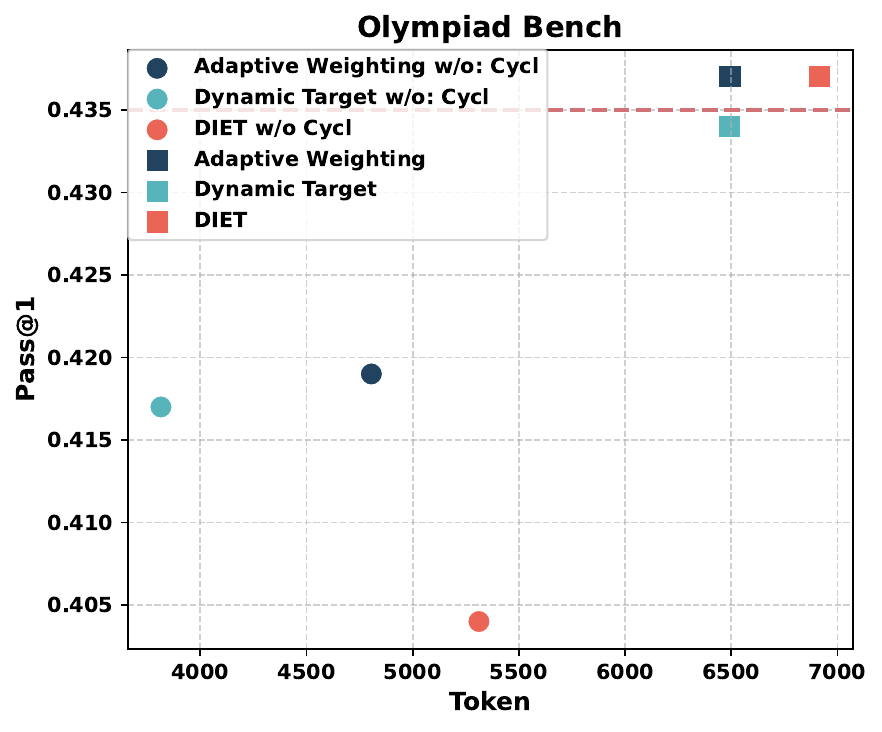}
  \end{minipage}%
  \begin{minipage}{0.33\textwidth}
    \centering
    \includegraphics[width=\linewidth]{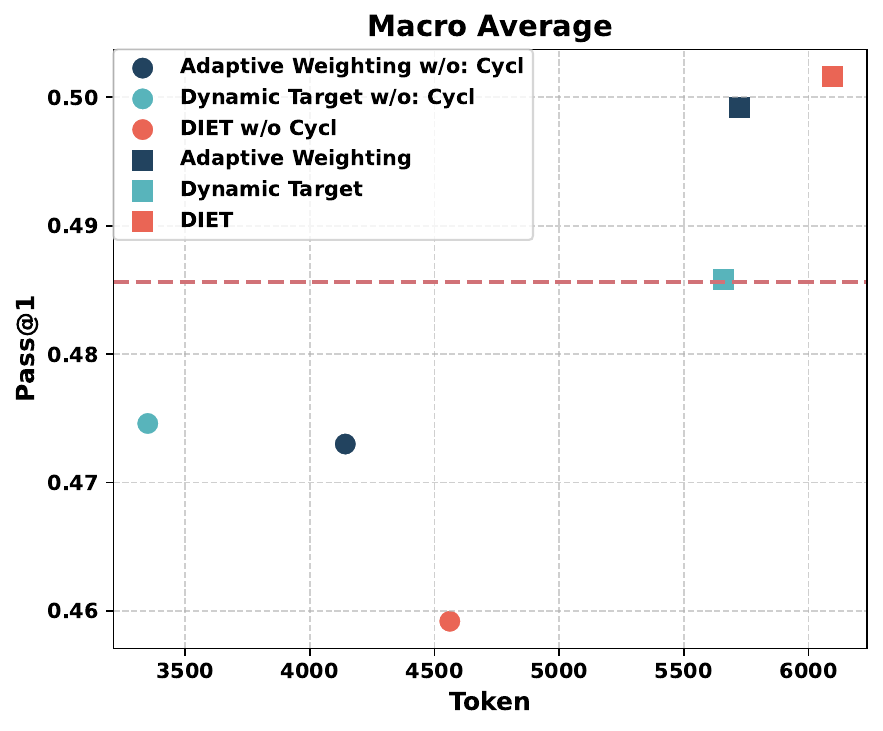}
  \end{minipage}
  
  \caption{Pass@1 versus average token count for Adaptive Weighting, Dynamic Target, and \framework configurations with and without cyclical compression, across various benchmarks and their Macro Average. The dashed red line indicates the Base Model's Pass@1 performance. Cyclical training generally trades a slight increase in tokens for improved Pass@1.}
  \label{fig:ablation_cyclical_impact}
\end{figure}

\cref{fig:ablation_cyclical_impact} illustrates these comparisons. Consistently across the benchmarks, particularly in the Macro Average (bottom right), applying cyclical compression shifts the methods to a higher Pass@1 compared to their non-cyclical counterparts. This performance improvement often allows the cyclical variants to meet or exceed the Base Model's Pass@1 (dashed red line).

This gain in reasoning accuracy typically corresponds to a moderate increase in average token length. For example, in the Macro Average plot, the \framework configuration achieves the highest Pass@1, while \framework w/o Cycl uses fewer tokens but results in the lowest Pass@1. This pattern suggests that while non-cyclical versions offer more aggressive token reduction, the "relax" phases in cyclical training are beneficial for achieving peak performance, justifying its inclusion in our best-performing \framework configurations presented in \cref{tab:reward-shaping-results-1.5b}. Thus, cyclical compression aids our difficulty-aware methods in achieving a superior balance of high performance and significant token savings relative to the Base Model.

\subsection{Impact of Trade-off Parameter \texorpdfstring{$\alpha_{base}$}{alpha}}
\label{app:ablation_alpha}
To determine an appropriate value for the trade-off parameter $\alpha_{base}$ in \cref{eq:rl-optim}, we conducted an ablation study on the Dynamic Target as a preliminary experiment. This preliminary experiment aimed to find a balance between maintaining reasoning performance and achieving significant token reduction. We tested $\alpha_{base} \in \{0.1, 0.5, 1.0\}$, with results shown in \cref{tab:ablation-alpha-1.5b}.

\begin{table}[t]
\centering
\caption{Macro Average Pass@1 (\%) and Token Count for the \textbf{Dynamic Target} method with varying trade-off parameter $\alpha_{base}$.}
\label{tab:ablation-alpha-1.5b}
\resizebox{1.0\textwidth}{!}{%
\begin{tabular}{l c c c c c c c c c c | c c}
\toprule
\multirow{2}{*}{\raisebox{-0.6\height}{\textbf{$\alpha_{base}$}}} & \multicolumn{2}{c}{\textbf{MATH 500}} & \multicolumn{2}{c}{\textbf{AIME 2024}} & \multicolumn{2}{c}{\textbf{AMC 2023}} & \multicolumn{2}{c}{\textbf{Olympiad.}} & \multicolumn{2}{c|}{\textbf{Minerva}} & \multicolumn{2}{c}{\textbf{Macro Average}} \\
\cmidrule(lr){2-3}\cmidrule(lr){4-5}\cmidrule(lr){6-7}\cmidrule(lr){8-9}\cmidrule(lr){10-11}\cmidrule(lr){12-13}
 & P@1 & Tok & P@1 & Tok & P@1 & Tok & P@1 & Tok & P@1 & Tok & P@1 & Tok \\ 
  \midrule
  0.1 & 83.1 & 2616 & 29.7 & 8852 & 65.1 & 5198 & 35.0 & 10326 & 27.1 & 2951 & 48.0 & 5988 \\
  0.5 & 80.1 & 1546 & 27.0 & 6794 & 63.6 & 3467 & 41.7 & 3814 & 24.9 & 1126 & 47.4 & 3349 \\
  1.0 & 73.5 & 1298 & 16.4 & 4217 & 49.5 & 1966 & 34.6 & 2475 & 22.2 & 1016 & 39.2 & 2194 \\
\bottomrule
\end{tabular}%
}
\end{table}

As shown in \cref{tab:ablation-alpha-1.5b}, increasing $\alpha_{base}$ leads to more aggressive token reduction but also a corresponding decrease in Pass@1 performance. Specifically, on Macro Average, increasing $\alpha_{base}$ from 0.1 to 1.0 reduces tokens significantly from 5988 to 2194, but Pass@1 drops from 48.0\% to 39.2\%. The setting of $\alpha_{base}=0.5$ yields a satisfactory trade-off. Based on this balance, we selected $\alpha_{base}=0.5$ as the default for all the difficulty-aware training in our main experiments.

\newpage
\section*{NeurIPS Paper Checklist}

\begin{enumerate}

\item {\bf Claims}
    \item[] Question: Do the main claims made in the abstract and introduction accurately reflect the paper's contributions and scope?
    \item[] Answer: \answerYes{} 
    \item[] Justification: We have outlined our main contributions in a point-by-point manner in both the abstract and the instruction sections. 
    \item[] Guidelines:
    \begin{itemize}
        \item The answer NA means that the abstract and introduction do not include the claims made in the paper.
        \item The abstract and/or introduction should clearly state the claims made, including the contributions made in the paper and important assumptions and limitations. A No or NA answer to this question will not be perceived well by the reviewers. 
        \item The claims made should match theoretical and experimental results, and reflect how much the results can be expected to generalize to other settings. 
        \item It is fine to include aspirational goals as motivation as long as it is clear that these goals are not attained by the paper. 
    \end{itemize}

\item {\bf Limitations}
    \item[] Question: Does the paper discuss the limitations of the work performed by the authors?
    \item[] Answer: \answerYes{} 
    \item[] Justification: In \cref{app:limitations}, we provide an analysis of the current limitations of our work. 
    \item[] Guidelines:
    \begin{itemize}
        \item The answer NA means that the paper has no limitation while the answer No means that the paper has limitations, but those are not discussed in the paper. 
        \item The authors are encouraged to create a separate "Limitations" section in their paper.
        \item The paper should point out any strong assumptions and how robust the results are to violations of these assumptions (e.g., independence assumptions, noiseless settings, model well-specification, asymptotic approximations only holding locally). The authors should reflect on how these assumptions might be violated in practice and what the implications would be.
        \item The authors should reflect on the scope of the claims made, e.g., if the approach was only tested on a few datasets or with a few runs. In general, empirical results often depend on implicit assumptions, which should be articulated.
        \item The authors should reflect on the factors that influence the performance of the approach. For example, a facial recognition algorithm may perform poorly when image resolution is low or images are taken in low lighting. Or a speech-to-text system might not be used reliably to provide closed captions for online lectures because it fails to handle technical jargon.
        \item The authors should discuss the computational efficiency of the proposed algorithms and how they scale with dataset size.
        \item If applicable, the authors should discuss possible limitations of their approach to address problems of privacy and fairness.
        \item While the authors might fear that complete honesty about limitations might be used by reviewers as grounds for rejection, a worse outcome might be that reviewers discover limitations that aren't acknowledged in the paper. The authors should use their best judgment and recognize that individual actions in favor of transparency play an important role in developing norms that preserve the integrity of the community. Reviewers will be specifically instructed to not penalize honesty concerning limitations.
    \end{itemize}

\item {\bf Theory assumptions and proofs}
    \item[] Question: For each theoretical result, does the paper provide the full set of assumptions and a complete (and correct) proof?
    \item[] Answer: \answerYes{} 
    \item[] Justification: We provide a derivation of advantage distortion under naive reward weighting in \cref{app:advantage_weighting_derivation} 
    \item[] Guidelines:
    \begin{itemize}
        \item The answer NA means that the paper does not include theoretical results. 
        \item All the theorems, formulas, and proofs in the paper should be numbered and cross-referenced.
        \item All assumptions should be clearly stated or referenced in the statement of any theorems.
        \item The proofs can either appear in the main paper or the supplemental material, but if they appear in the supplemental material, the authors are encouraged to provide a short proof sketch to provide intuition. 
        \item Inversely, any informal proof provided in the core of the paper should be complemented by formal proofs provided in appendix or supplemental material.
        \item Theorems and Lemmas that the proof relies upon should be properly referenced. 
    \end{itemize}

    \item {\bf Experimental result reproducibility}
    \item[] Question: Does the paper fully disclose all the information needed to reproduce the main experimental results of the paper to the extent that it affects the main claims and/or conclusions of the paper (regardless of whether the code and data are provided or not)?
    \item[] Answer: \answerYes{} 
    \item[] Justification: We provide test details in \cref{app:eval_details}. We will also open-source our evaluation code after proper organization.
    \item[] Guidelines:
    \begin{itemize}
        \item The answer NA means that the paper does not include experiments.
        \item If the paper includes experiments, a No answer to this question will not be perceived well by the reviewers: Making the paper reproducible is important, regardless of whether the code and data are provided or not.
        \item If the contribution is a dataset and/or model, the authors should describe the steps taken to make their results reproducible or verifiable. 
        \item Depending on the contribution, reproducibility can be accomplished in various ways. For example, if the contribution is a novel architecture, describing the architecture fully might suffice, or if the contribution is a specific model and empirical evaluation, it may be necessary to either make it possible for others to replicate the model with the same dataset, or provide access to the model. In general. releasing code and data is often one good way to accomplish this, but reproducibility can also be provided via detailed instructions for how to replicate the results, access to a hosted model (e.g., in the case of a large language model), releasing of a model checkpoint, or other means that are appropriate to the research performed.
        \item While NeurIPS does not require releasing code, the conference does require all submissions to provide some reasonable avenue for reproducibility, which may depend on the nature of the contribution. For example
        \begin{enumerate}
            \item If the contribution is primarily a new algorithm, the paper should make it clear how to reproduce that algorithm.
            \item If the contribution is primarily a new model architecture, the paper should describe the architecture clearly and fully.
            \item If the contribution is a new model (e.g., a large language model), then there should either be a way to access this model for reproducing the results or a way to reproduce the model (e.g., with an open-source dataset or instructions for how to construct the dataset).
            \item We recognize that reproducibility may be tricky in some cases, in which case authors are welcome to describe the particular way they provide for reproducibility. In the case of closed-source models, it may be that access to the model is limited in some way (e.g., to registered users), but it should be possible for other researchers to have some path to reproducing or verifying the results.
        \end{enumerate}
    \end{itemize}

\item {\bf Open access to data and code}
    \item[] Question: Does the paper provide open access to the data and code, with sufficient instructions to faithfully reproduce the main experimental results, as described in supplemental material?
    \item[] Answer: \answerNo{} 
    \item[] Justification: We will open-source our code after proper organization. 
    \item[] Guidelines:
    \begin{itemize}
        \item The answer NA means that paper does not include experiments requiring code.
        \item Please see the NeurIPS code and data submission guidelines (\url{https://nips.cc/public/guides/CodeSubmissionPolicy}) for more details.
        \item While we encourage the release of code and data, we understand that this might not be possible, so “No” is an acceptable answer. Papers cannot be rejected simply for not including code, unless this is central to the contribution (e.g., for a new open-source benchmark).
        \item The instructions should contain the exact command and environment needed to run to reproduce the results. See the NeurIPS code and data submission guidelines (\url{https://nips.cc/public/guides/CodeSubmissionPolicy}) for more details.
        \item The authors should provide instructions on data access and preparation, including how to access the raw data, preprocessed data, intermediate data, and generated data, etc.
        \item The authors should provide scripts to reproduce all experimental results for the new proposed method and baselines. If only a subset of experiments are reproducible, they should state which ones are omitted from the script and why.
        \item At submission time, to preserve anonymity, the authors should release anonymized versions (if applicable).
        \item Providing as much information as possible in supplemental material (appended to the paper) is recommended, but including URLs to data and code is permitted.
    \end{itemize}

\item {\bf Experimental setting/details}
    \item[] Question: Does the paper specify all the training and test details (e.g., data splits, hyperparameters, how they were chosen, type of optimizer, etc.) necessary to understand the results?
    \item[] Answer: \answerYes{} 
    \item[] Justification: We provide detailed training details in 
\cref{app:training_details} and detailed evaluation details in \cref{app:eval_details}. 
    \item[] Guidelines:
    \begin{itemize}
        \item The answer NA means that the paper does not include experiments.
        \item The experimental setting should be presented in the core of the paper to a level of detail that is necessary to appreciate the results and make sense of them.
        \item The full details can be provided either with the code, in appendix, or as supplemental material.
    \end{itemize}

\item {\bf Experiment statistical significance}
    \item[] Question: Does the paper report error bars suitably and correctly defined or other appropriate information about the statistical significance of the experiments?
    \item[] Answer: \answerYes{} 
    \item[] Justification: We conducted a p-value test when computing the Pearson correlation between problem difficulty and average response length and ensured that the p-value is less than 0.01. And we reported error bars in  \cref{fig:base-corr} .
    \item[] Guidelines:
    \begin{itemize}
        \item The answer NA means that the paper does not include experiments.
        \item The authors should answer "Yes" if the results are accompanied by error bars, confidence intervals, or statistical significance tests, at least for the experiments that support the main claims of the paper.
        \item The factors of variability that the error bars are capturing should be clearly stated (for example, train/test split, initialization, random drawing of some parameter, or overall run with given experimental conditions).
        \item The method for calculating the error bars should be explained (closed form formula, call to a library function, bootstrap, etc.)
        \item The assumptions made should be given (e.g., Normally distributed errors).
        \item It should be clear whether the error bar is the standard deviation or the standard error of the mean.
        \item It is OK to report 1-sigma error bars, but one should state it. The authors should preferably report a 2-sigma error bar than state that they have a 96\% CI, if the hypothesis of Normality of errors is not verified.
        \item For asymmetric distributions, the authors should be careful not to show in tables or figures symmetric error bars that would yield results that are out of range (e.g. negative error rates).
        \item If error bars are reported in tables or plots, The authors should explain in the text how they were calculated and reference the corresponding figures or tables in the text.
    \end{itemize}

\item {\bf Experiments compute resources}
    \item[] Question: For each experiment, does the paper provide sufficient information on the computer resources (type of compute workers, memory, time of execution) needed to reproduce the experiments?
    \item[] Answer: \answerYes{} 
    \item[] Justification: We detail the computational resources used in our experiments in \cref{sec:exp_setup}. 
    \item[] Guidelines:
    \begin{itemize}
        \item The answer NA means that the paper does not include experiments.
        \item The paper should indicate the type of compute workers CPU or GPU, internal cluster, or cloud provider, including relevant memory and storage.
        \item The paper should provide the amount of compute required for each of the individual experimental runs as well as estimate the total compute. 
        \item The paper should disclose whether the full research project required more compute than the experiments reported in the paper (e.g., preliminary or failed experiments that didn't make it into the paper). 
    \end{itemize}
    
\item {\bf Code of ethics}
    \item[] Question: Does the research conducted in the paper conform, in every respect, with the NeurIPS Code of Ethics \url{https://neurips.cc/public/EthicsGuidelines}?
    \item[] Answer: \answerYes{} 
    \item[] Justification: Our research adheres to the NeurIPS Code of Ethics.  It does not involve human subjects, personally identifiable information, or sensitive data.  
    \item[] Guidelines:
    \begin{itemize}
        \item The answer NA means that the authors have not reviewed the NeurIPS Code of Ethics.
        \item If the authors answer No, they should explain the special circumstances that require a deviation from the Code of Ethics.
        \item The authors should make sure to preserve anonymity (e.g., if there is a special consideration due to laws or regulations in their jurisdiction).
    \end{itemize}

\item {\bf Broader impacts}
    \item[] Question: Does the paper discuss both potential positive societal impacts and negative societal impacts of the work performed?
    \item[] Answer: \answerNA{} 
    \item[] Justification: Our experiments are conducted using the already widely-used open-source model, which will not induce new societal impact.
    \item[] Guidelines:
    \begin{itemize}
        \item The answer NA means that there is no societal impact of the work performed.
        \item If the authors answer NA or No, they should explain why their work has no societal impact or why the paper does not address societal impact.
        \item Examples of negative societal impacts include potential malicious or unintended uses (e.g., disinformation, generating fake profiles, surveillance), fairness considerations (e.g., deployment of technologies that could make decisions that unfairly impact specific groups), privacy considerations, and security considerations.
        \item The conference expects that many papers will be foundational research and not tied to particular applications, let alone deployments. However, if there is a direct path to any negative applications, the authors should point it out. For example, it is legitimate to point out that an improvement in the quality of generative models could be used to generate deepfakes for disinformation. On the other hand, it is not needed to point out that a generic algorithm for optimizing neural networks could enable people to train models that generate Deepfakes faster.
        \item The authors should consider possible harms that could arise when the technology is being used as intended and functioning correctly, harms that could arise when the technology is being used as intended but gives incorrect results, and harms following from (intentional or unintentional) misuse of the technology.
        \item If there are negative societal impacts, the authors could also discuss possible mitigation strategies (e.g., gated release of models, providing defenses in addition to attacks, mechanisms for monitoring misuse, mechanisms to monitor how a system learns from feedback over time, improving the efficiency and accessibility of ML).
    \end{itemize}
    
\item {\bf Safeguards}
    \item[] Question: Does the paper describe safeguards that have been put in place for responsible release of data or models that have a high risk for misuse (e.g., pretrained language models, image generators, or scraped datasets)?
    \item[] Answer: \answerNA{} 
    \item[] Justification: We use open-source reasoning model for our experiments, which will pose no such risks. 
    \item[] Guidelines:
    \begin{itemize}
        \item The answer NA means that the paper poses no such risks.
        \item Released models that have a high risk for misuse or dual-use should be released with necessary safeguards to allow for controlled use of the model, for example by requiring that users adhere to usage guidelines or restrictions to access the model or implementing safety filters. 
        \item Datasets that have been scraped from the Internet could pose safety risks. The authors should describe how they avoided releasing unsafe images.
        \item We recognize that providing effective safeguards is challenging, and many papers do not require this, but we encourage authors to take this into account and make a best faith effort.
    \end{itemize}

\item {\bf Licenses for existing assets}
    \item[] Question: Are the creators or original owners of assets (e.g., code, data, models), used in the paper, properly credited and are the license and terms of use explicitly mentioned and properly respected?
    \item[] Answer: \answerYes{} 
    \item[] Justification: All external assets used in our work, including code(e.g., verl, sympy), datasets(e.g., deepscaler), and models(e.g. R1-Distilled Qwen), are properly credited. We ensure that the licenses and terms of use for these resources are explicitly stated and strictly followed.  
    \item[] Guidelines:
    \begin{itemize}
        \item The answer NA means that the paper does not use existing assets.
        \item The authors should cite the original paper that produced the code package or dataset.
        \item The authors should state which version of the asset is used and, if possible, include a URL.
        \item The name of the license (e.g., CC-BY 4.0) should be included for each asset.
        \item For scraped data from a particular source (e.g., website), the copyright and terms of service of that source should be provided.
        \item If assets are released, the license, copyright information, and terms of use in the package should be provided. For popular datasets, \url{paperswithcode.com/datasets} has curated licenses for some datasets. Their licensing guide can help determine the license of a dataset.
        \item For existing datasets that are re-packaged, both the original license and the license of the derived asset (if it has changed) should be provided.
        \item If this information is not available online, the authors are encouraged to reach out to the asset's creators.
    \end{itemize}

\item {\bf New assets}
    \item[] Question: Are new assets introduced in the paper well documented and is the documentation provided alongside the assets?
    \item[] Answer: \answerYes{} 
    \item[] Justification: We will open-source our code and provide detailed documentation and comments for ease of use. 
    \item[] Guidelines:
    \begin{itemize}
        \item The answer NA means that the paper does not release new assets.
        \item Researchers should communicate the details of the dataset/code/model as part of their submissions via structured templates. This includes details about training, license, limitations, etc. 
        \item The paper should discuss whether and how consent was obtained from people whose asset is used.
        \item At submission time, remember to anonymize your assets (if applicable). You can either create an anonymized URL or include an anonymized zip file.
    \end{itemize}

\item {\bf Crowdsourcing and research with human subjects}
    \item[] Question: For crowdsourcing experiments and research with human subjects, does the paper include the full text of instructions given to participants and screenshots, if applicable, as well as details about compensation (if any)? 
    \item[] Answer: \answerNA{} 
    \item[] Justification: Crowdsourcing and human subjects are not involved in our research. 
    \item[] Guidelines:
    \begin{itemize}
        \item The answer NA means that the paper does not involve crowdsourcing nor research with human subjects.
        \item Including this information in the supplemental material is fine, but if the main contribution of the paper involves human subjects, then as much detail as possible should be included in the main paper. 
        \item According to the NeurIPS Code of Ethics, workers involved in data collection, curation, or other labor should be paid at least the minimum wage in the country of the data collector. 
    \end{itemize}

\item {\bf Institutional review board (IRB) approvals or equivalent for research with human subjects}
    \item[] Question: Does the paper describe potential risks incurred by study participants, whether such risks were disclosed to the subjects, and whether Institutional Review Board (IRB) approvals (or an equivalent approval/review based on the requirements of your country or institution) were obtained?
    \item[] Answer: \answerNA{} 
    \item[] Justification: Crowdsourcing and human subjects are not involved in our research. 
    \item[] Guidelines:
    \begin{itemize}
        \item The answer NA means that the paper does not involve crowdsourcing nor research with human subjects.
        \item Depending on the country in which research is conducted, IRB approval (or equivalent) may be required for any human subjects research. If you obtained IRB approval, you should clearly state this in the paper. 
        \item We recognize that the procedures for this may vary significantly between institutions and locations, and we expect authors to adhere to the NeurIPS Code of Ethics and the guidelines for their institution. 
        \item For initial submissions, do not include any information that would break anonymity (if applicable), such as the institution conducting the review.
    \end{itemize}

\item {\bf Declaration of LLM usage}
    \item[] Question: Does the paper describe the usage of LLMs if it is an important, original, or non-standard component of the core methods in this research? Note that if the LLM is used only for writing, editing, or formatting purposes and does not impact the core methodology, scientific rigorousness, or originality of the research, declaration is not required.
    \item[] Answer: \answerNA{} 
    \item[] Justification: We only used LLM for writing and editing which does not impact the core methodology. 
    \item[] Guidelines:
    \begin{itemize}
        \item The answer NA means that the core method development in this research does not involve LLMs as any important, original, or non-standard components.
        \item Please refer to our LLM policy (\url{https://neurips.cc/Conferences/2025/LLM}) for what should or should not be described.
    \end{itemize}

\end{enumerate}

\end{document}